\documentclass[acmsmall, screen]{acmart}

\AtBeginDocument{%
  }

\acmJournal{CSUR}
\acmVolume{37}
\acmNumber{4}
\acmArticle{111}
\acmMonth{8}

\usepackage{subfig}
\usepackage{graphicx}
\usepackage{multirow}
\usepackage{pifont}
\usepackage{pax}
\newcommand{\cmark}{\ding{51}}%

\newcommand{\etal}{\textit{et al}.}

\newcommand{\eg}{\textit{e}.\textit{g}.}

\begin{document}

\title{Deepfake Media Generation and Detection in the Generative AI Era: A Survey and Outlook}

\author{Florinel Alin Croitoru}
\email{alincroitoru97@gmail.com}
\authornote{All authors contributed equally to this research.}
\affiliation{%
  \institution{Department of Computer Science, University of Bucharest}
  \city{Bucharest}
  \country{Romania}
}

\author{Andrei-Iulian  H\^{i}ji}
\email{andrei-iulian.hiji@unibuc.ro}
\authornotemark[1]
\affiliation{%
  \institution{Department of Computer Science, University of Bucharest}
  \city{Bucharest}
  \country{Romania}
}

\author{Vlad Hondru}
\email{vlad.hondru@fmi.unibuc.ro}
\authornotemark[1]
\affiliation{%
  \institution{Department of Computer Science, University of Bucharest}
  \city{Bucharest}
  \country{Romania}
}

\author{Nicolae C\u{a}t\u{a}lin Ristea}
\email{ristea.nc@gmail.com}
\authornotemark[1]
\affiliation{%
  \institution{Department of Computer Science, University of Bucharest}
  \city{Bucharest}
  \country{Romania}
}

\author{Paul Irofti}
\email{paul@irofti.net}
\affiliation{%
  \institution{Department of Computer Science, University of Bucharest}
  \city{Bucharest}
  \country{Romania}
}

\author{Marius Popescu}
\email{popescunmarius@gmail.com}
\affiliation{%
  \institution{Department of Computer Science, University of Bucharest}
  \city{Bucharest}
  \country{Romania}
}

\author{Cristian Rusu}
\email{cristian.rusu@fmi.unibuc.ro}
\affiliation{%
  \institution{Department of Computer Science, University of Bucharest}
  \city{Bucharest}
  \country{Romania}
}

\author{Radu-Tudor Ionescu}
\email{raducu.ionescu@gmail.com}
\authornote{Corresponding author.}
\affiliation{%
  \institution{Department of Computer Science, University of Bucharest}
  \city{Bucharest}
  \country{Romania}
}

\author{Fahad-Shahbaz Khan}
\email{fahad.khan@liu.se}
\affiliation{%
  \institution{Mohamed bin Zayed University of Artificial Intelligence (MBZUAI)}
  \city{Abu Dhabi}
  \country{UAE}
}

\author{Mubarak Shah}
\email{shah@crcv.ucf.edu}
\affiliation{%
  \institution{Department of Computer Science, University of Central Florida}
  \city{Orlando}
  \country{US}
}

\renewcommand{\shortauthors}{Croitoru et al.}

\begin{abstract}
We survey deepfake generation and detection techniques, covering all deepfake media types: image, video, audio and multimodal content. We identify various kinds of deepfakes and construct taxonomies of deepfake generation and detection methods, illustrating the important groups of methods. Next, we gather datasets used for deepfake detection and provide updated rankings of the best performing  detectors on the most popular datasets. In addition, we develop a novel multimodal benchmark to evaluate deepfake detectors on out-of-distribution content. The results indicate that state-of-the-art detectors fail to generalize to deepfakes generated by unseen generators. Our project page and new benchmark are available at \url{https://github.com/CroitoruAlin/biodeep}.
\end{abstract}



\received{20 February 2007}
\received[revised]{12 March 2009}
\received[accepted]{5 June 2009}

\maketitle

\setlength{\abovedisplayskip}{3.5pt}
\setlength{\belowdisplayskip}{3.5pt}

\section{Introduction}\label{sec:introduction}
Deepfake media comprises image, video or audio files that are digitally altered or generated from scratch with AI tools in order to impersonate real or non-existent people. The recent groundbreaking progress of generative AI methods \cite{Croitoru-TPAMI-2023, Chen-ICLR-2024, Rombach-CVPR-2022, Podell-ICLR-2024, Sauer-SIGGRAPH-2022, Esser-CVPR-2021} has enabled the creation of realistic deepfake media with very little effort \cite{Li-CVPR-2023, Liu-CVPR-2023, Kim-CVPR-2022, Shiohara-CVPR-2023, Papantoniou-ECCV-2024, Zhao-CVPR-2023, Liu-CVPR-2024, nirkin-ICCV-2019, Bounareli-ICCV-2023, oorloff-ICCV-2023, Ma-ArXiv-2023, Xu-ArXiv-2024a}. Unfortunately, the generated deepfake media can be used by scammers to spread misinformation on social media platforms to achieve large-scale political manipulation, and to deceive individuals or companies into financial frauds.

In an age where misinformation can quickly spread through social media platforms, deepfakes pose a critical threat to public trust and democracy, especially due to their growing online exploitation. A recent analysis of the fraud trends indicates that the number of fraud cases based on deepfakes registered a 10$\times$ increase in 2023, with respect to 2022\footnote{\href{https://sumsub.com/blog/sumsub-experts-top-kyc-trends-2024/}{Sumsub Expert Roundtable: The Top KYC Trends Coming in 2024}}. Another recent study found that about $70\%$ of people are unable to distinguish between a real and a deepfake voice\footnote{\href{https://www.mcafee.com/blogs/privacy-identity-protection/artificial-imposters-cybercriminals-turn-to-ai-voice-cloning-for-a-new-breed-of-scam/}{Artificial Imposters--Cybercriminals Turn to AI Voice Cloning for a New Breed of Scam}}. The growing quality and quantity of deepfakes raise significant concerns, particularly regarding online fraud and manipulation. To prevent the spread of deepfake media, researchers have developed a broad range of unimodal \cite{Shiohara-CVPR-2022, Chen-CVPR-2022, Dong-CVPR-2023, Lin-CVPR-2024b, Xu-IJCV-2024} or multimodal \cite{Zhou-ICCV-2021, Cozzolino-CVPR-2023, Goyal-TCSS-2023} methods for deepfake detection. However, deepfake detectors trained on media generated with a certain set of AI tools typically fail on deepfakes generated with a distinct set of tools \cite{Chen-CVPR-2022, Dong-CVPR-2023, Lin-CVPR-2024b}. This has led to a relentless race to develop more powerful and robust deepfake detectors.

To this end, we conduct a comprehensive survey on the recent developments in deepfake media generation and detection, focusing on synthetic content designed to impersonate real individuals. We first define a set of deepfake categories, which are determined based on the procedure used to generate the deepfake content. We identify both domain-agnostic and domain-specific deepfake types, and explain what kind of deepfake media belongs to each category. We next build taxonomies of deepfake generation and detection methods, which create multi-perspective hierarchical categorizations based on the considered media types, the employed architectures and the targeted tasks. 
As per Figures A1 and A2 (in supplementary), we first divide contributions by task, into generation and detection. For each task, we identify the employed architectures. For deepfake generation, we find that the most popular methods are based on Generative Adversarial Networks (GANs) \cite{Natsume-SIGGRAPH-2018, nirkin-ICCV-2019, skorokhodov-CVPR-2022, oorloff-ICCV-2023, Liu-CVPR-2023, Bounareli-ICCV-2023}, denoising diffusion models \cite{Papantoniou-ECCV-2024, Zhao-CVPR-2023, Liu-CVPR-2024, Kim-ArXiv-2024, Wang-CVPR-2024,  Xu-ArXiv-2024a, Stypulkowski-WACV-2024}, transformers~\cite{Jang-CVPR-2024, Cheng-SIGGRAPH-2022, Wang-AAAI-2022, Ling-JSTSP-2023, Zhao-CVPR-2025}, Gaussian splatting~\cite{Wang-TVCG-2025, Cho-MM-2024, Li-CVPR-2025} or NeRF~\cite{Peng-CVPR-2024a, Ye-ArXiv-2023}. 
To detect deepfakes, the majority of methods are based on convolutional neural networks (CNNs) \cite{Zhou-ICCV-2021, Cozzolino-CVPR-2023, Shiohara-CVPR-2022, Dong-CVPR-2023}, transformers \cite{Wang-ICMR-2022, Dong-CVPR-2022, cai-CVPR-2023}, or hybrid architectures that combine CNNs either with transformers \cite{zheng-ICCV-2021, choi-CVPR-2024, Shao-ECCV-2022} or recurrent neural networks (RNNs) \cite{guera-AVSS-2018, Liu-WACV-2023}. For each type of architecture, we further divide the contributions with respect to the media types: image, video, audio or multimodal (audio-visual). 

After presenting the main contributions in each category of articles included in the taxonomies, we review existing datasets for deepfake detection in image, video and audio. We then aggregate the reported performance levels of deepfake detectors on the most popular datasets, thus facilitating a direct comparison of existing methods. In addition, we introduce a benchmark to test the generalization capacity of deepfake detectors to out-of-distribution content. Interestingly, we find that state-of-the-art deepfake detectors showcase poor generalization to realistic deepfakes generated by newer and more powerful generative models. Finally, we identify research gaps in current literature, proposing a series of future work directions that can lead to the development of better frameworks to detect deepfake media. Among the proposed future directions, we promote a relatively new task formulation for deepfake detection, called person-of-interest (POI) deepfake detection, where each test sample comes with a verified (real) instance of the target person, thereby reducing the difficulty of the detection task. Another proposal is the development of a blockchain solution for multimedia sharing, which aims to provide a mechanism to only share verified content and to precisely determine the source of shared multimedia content, thereby eliminating the need to perform deepfake detection.


In summary, our contribution is fourfold:
\begin{itemize}
    \item We conduct a comprehensive survey of deepfake generation and detection methods, comprising recent advancements in four domains: image, video, audio and multimodal.
    \item We construct a taxonomy of deepfake generation and detection methods, categorizing research articles according to tasks, architectures and media types.
    \item We collect and merge results reported on popular deepfake detection benchmarks, providing the means to easily assess the current performance levels of deepfake detectors.
    \item We introduce a benchmark to test the out-of-domain generalization of deepfake detection models, showing that current detectors generally exhibit high performance drops on deepfakes generated by new and powerful generators.
\end{itemize}

\section{Related Surveys}

\begin{table}[t]
\caption{Comparing our survey with related surveys in terms of the covered tasks, domains, methods and other aspects. There are at least six factors that differentiate our survey from each of the other surveys.}
\label{table:surveys}  
\vspace{-0.3cm}
\scriptsize{
\centering
\begin{tabular}{l|ccc|cccc|cccccc|ccccc}
 \toprule 
 \multirow{7.5}{*}{Survey}   &   \multicolumn{3}{c|}{Task}     &  \multicolumn{4}{c|}{Domain} & \multicolumn{6}{c|}{Method} & & & \\
\cline{2-14}
 & \rotatebox{90}{Generation}  &   \rotatebox{90}{Detection}  & \rotatebox{90}{POI Detection} &  \rotatebox{90}{Image}  &   \rotatebox{90}{Video}    &  \rotatebox{90}{Audio}   &  \rotatebox{90}{Multimodal}   & \rotatebox{90}{GANs} & \rotatebox{90}{Diffusion} & \rotatebox{90}{CNNs} & \rotatebox{90}{RNNs} & \rotatebox{90}{Transformers} & \rotatebox{90}{Others} & \rotatebox{90}{Taxonomies} & \rotatebox{90}{Datasets} & \rotatebox{90}{Tutorial} & \rotatebox{90}{New Benchmark} & \rotatebox{90}{Blockchain Solution$\!\!\!\!\!$} \\
\midrule
Das \etal~\cite{Das-ICNGIS-2022} &  & \cmark & &  & \cmark &  &  &  &  & \cmark & \cmark & & & & & & & \\
\hline
Heidari \etal~\cite{Heidari-WIREs-2024} & & \cmark & & \cmark & \cmark & \cmark & \cmark &  &  & \cmark & \cmark & & \cmark & & & & & \\
\hline
Kaur \etal~\cite{Kaur-AIR-2024} & \cmark & \cmark & &  & \cmark &  &  & \cmark & \cmark & \cmark & \cmark & \cmark & \cmark & \cmark & \cmark &  & & \\
\hline
Lei \etal~\cite{Lei-ArXiv-2024} & \cmark &  & &  & \cmark &  &  & \cmark & \cmark &  &  &  & \cmark & & \cmark & & & \\
\hline
Li \etal~\cite{Li-ArXiv-2024b} & & \cmark & & & & \cmark & & & & \cmark & \cmark & \cmark & \cmark & \cmark & \cmark & & & \\
\hline
Li \etal~\cite{Li-ArXiv-2024} & \cmark &  & &  & \cmark &  &  & \cmark & \cmark & \cmark &  & \cmark & \cmark & & \cmark & & & \\
\hline
Masood \etal~\cite{Masood-AI-2023} & \cmark & \cmark & & \cmark & \cmark & \cmark & \cmark & \cmark &  & \cmark & \cmark & & \cmark & \cmark & \cmark & & & \\
\hline
Patel \etal~\cite{Patel-Access-2023} & \cmark & \cmark & & \cmark & \cmark & \cmark & \cmark & \cmark &  & \cmark & \cmark & & \cmark & \cmark & \cmark & & & \\
\hline
Pei \etal~\cite{Pei-ArXiv-2024} & \cmark & \cmark & & \cmark & \cmark & & & \cmark & \cmark & \cmark & \cmark & \cmark & \cmark & \cmark & \cmark & & & \\
\hline
Seow \etal~\cite{Seow-NEUCOM-2022} & \cmark & \cmark & & \cmark & \cmark &  &  & \cmark &  & \cmark & \cmark & & \cmark & & \cmark & & & \\
\hline
Yi \etal~\cite{Yi-ArXiv-2023} &  & \cmark & &  &  & \cmark &  &  &  & \cmark & \cmark & \cmark & \cmark & \cmark & \cmark & & & \\
\hline
Zhang~\cite{Zhang-MTA-2022} & \cmark & \cmark & & \cmark & \cmark & \cmark &  & \cmark &  & \cmark & \cmark & & \cmark &  & \cmark & & & \\
\hline
Ours & \cmark & \cmark & \cmark & \cmark & \cmark & \cmark & \cmark & \cmark & \cmark & \cmark & \cmark & \cmark & \cmark & \cmark & \cmark & \cmark & \cmark  & \cmark \\
\bottomrule
\end{tabular}

}
\end{table}

Several attempts have been made to survey deepfake detection and generation. In Table \ref{table:surveys}, we gather related surveys and illustrate the tasks, domains, methods and other aspects covered by the gathered surveys. Some surveys only cover the generation part \cite{Lei-ArXiv-2024,Li-ArXiv-2024}, while others are particularly focused on detection \cite{Das-ICNGIS-2022,Heidari-WIREs-2024,Li-ArXiv-2024b,Yi-ArXiv-2023}. Many surveys consider only one input media type, \eg~video \cite{Das-ICNGIS-2022,Kaur-AIR-2024,Lei-ArXiv-2024,Li-ArXiv-2024} or audio \cite{Li-ArXiv-2024b,Yi-ArXiv-2023}. There are a few surveys \cite{Heidari-WIREs-2024,Masood-AI-2023,Patel-Access-2023} that cover all media types (image, video, audio and multimodal), but only Masood \etal~\cite{Masood-AI-2023} and Patel \etal~\cite{Patel-Access-2023} address both detection and generation tasks. Although the surveys of Masood \etal~\cite{Masood-AI-2023} and Patel \etal~\cite{Patel-Access-2023} are comprehensive, they do not cover the most recent developments, such as diffusion models and vision transformers. 

In summary, we find that existing surveys are either outdated or limited in terms of coverage, including only specific tasks (generation or detection) or media types (image, audio or video). In contrast, we conduct \emph{an extensive survey of current literature}, covering both generation and detection, as well as all deepfake media types. We also discuss a novel task formulation, namely POI deepfake detection, which aims to simplify the detection task. Moreover, we create multi-level taxonomies for generation and detection to ease the navigation through the current deepfake literature, providing direct links to the referenced papers. The taxonomy is complemented by a tutorial comprising four of the most predominant generative frameworks. To our knowledge, our survey is the first to propose \emph{a novel benchmark} to test the generalization capacity of deepfake detectors to out-of-distribution data. Another aspect that distinguishes our survey from the rest is the proposal of blockchain solution for multimedia sharing.

\begin{figure}[!t]
    \centering
        \subfloat[Identity swapping.\vspace{-0.15cm}]{\includegraphics[width=0.315\linewidth]{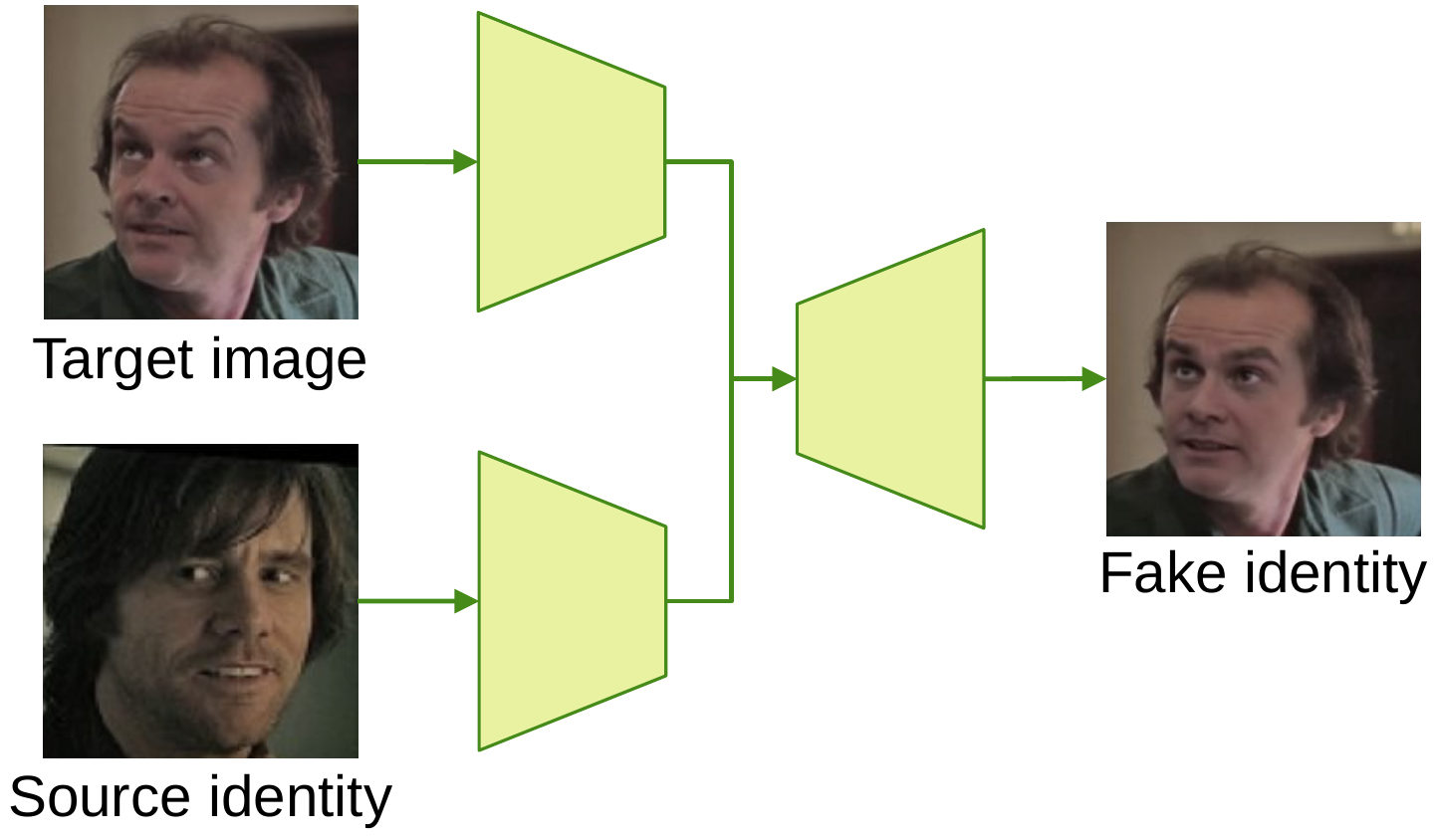}}
        \hfill
        \subfloat[Facial expression swapping.\vspace{-0.15cm}]{\includegraphics[width=0.33\linewidth]{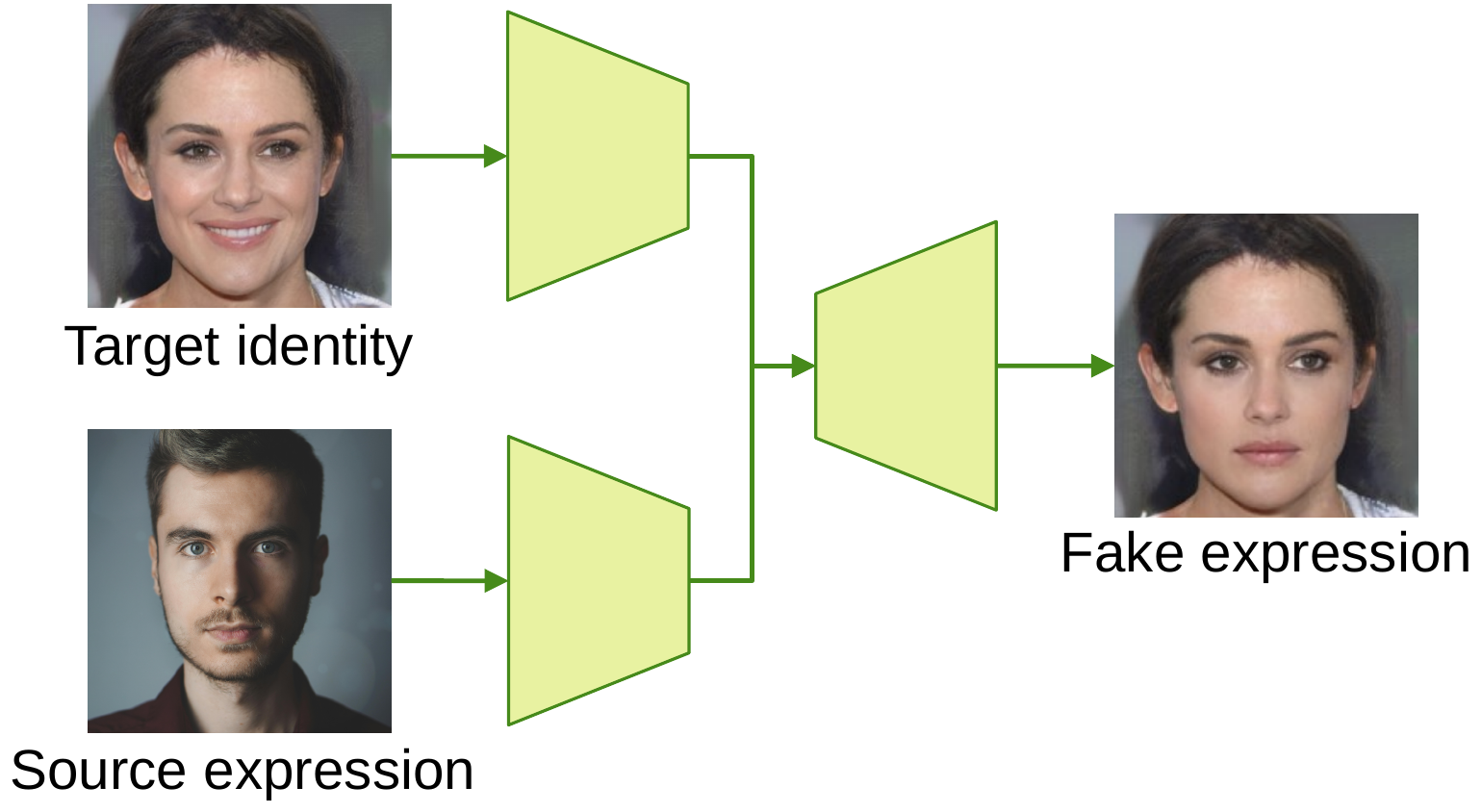}}
        \hfill
        \subfloat[Facial attribute manipulation.\vspace{-0.15cm}]{\includegraphics[width=0.3\linewidth]{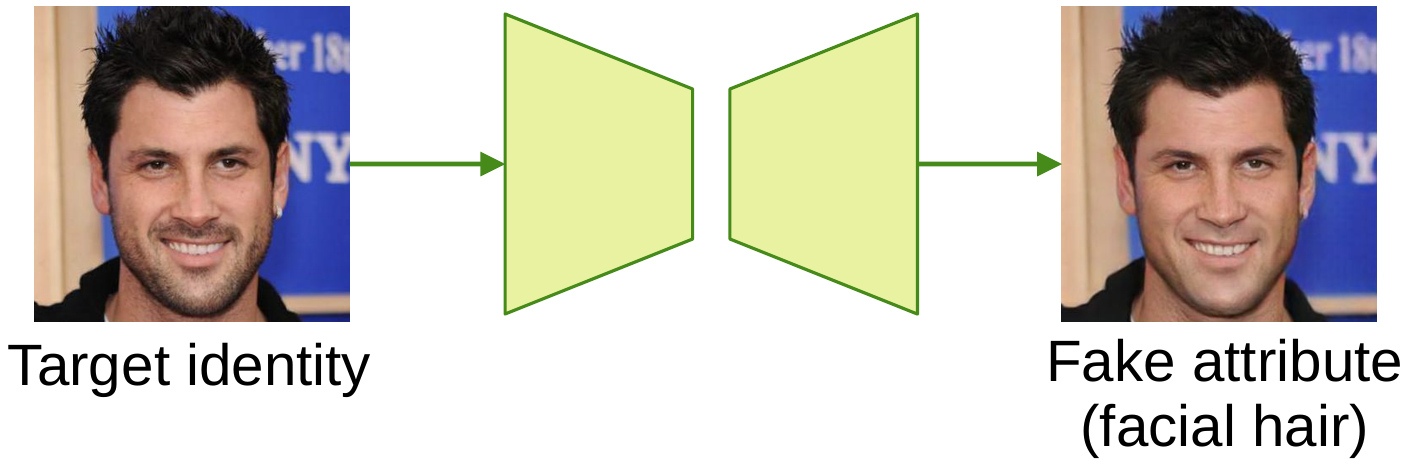}}
        \\
         \subfloat[Talking face synthesis.\vspace{-0.15cm}]{\includegraphics[width=0.305\linewidth]{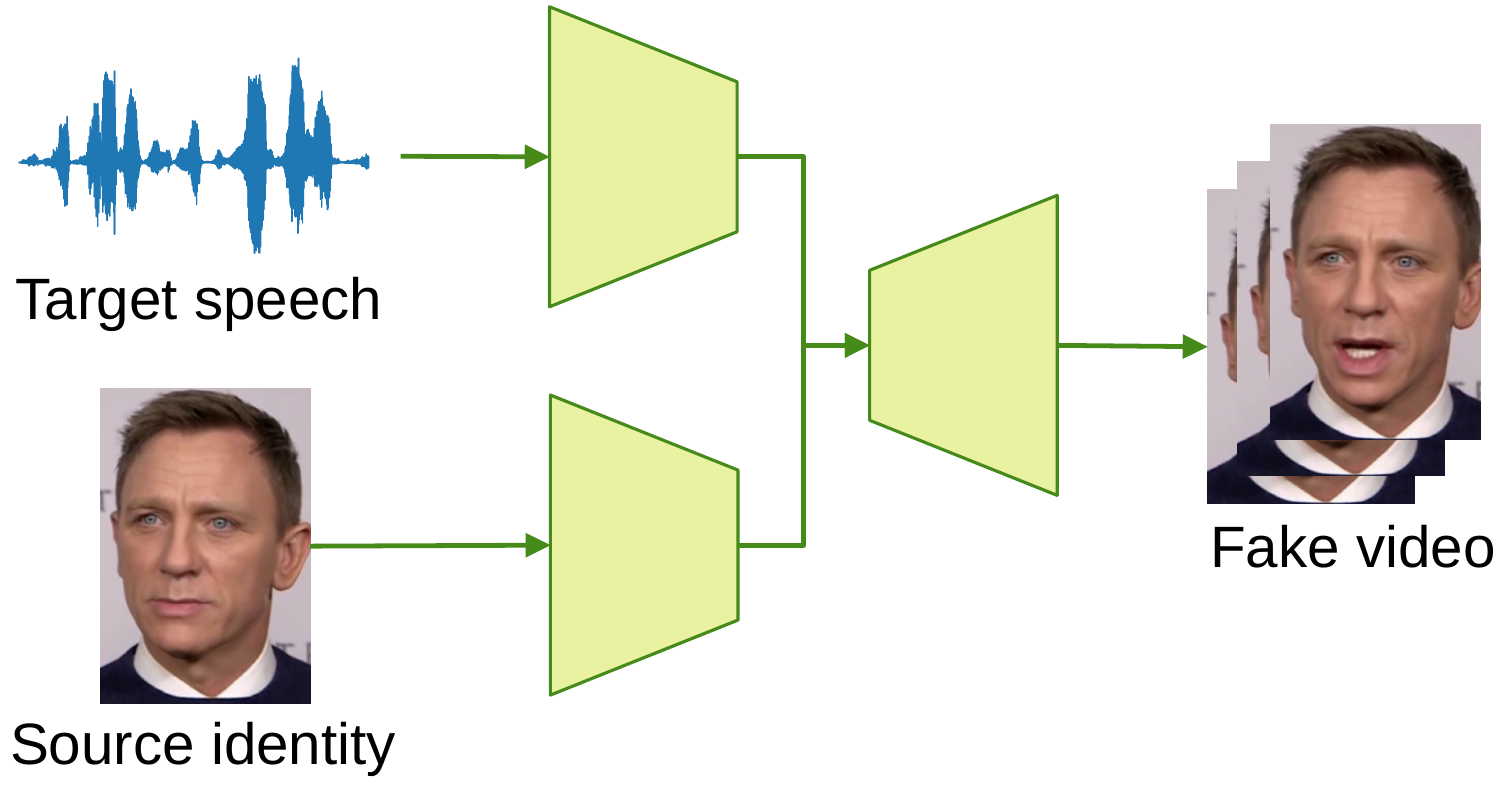}}
        \hfill
        \subfloat[Background swapping.\vspace{-0.15cm}]{\includegraphics[width=0.315\linewidth]{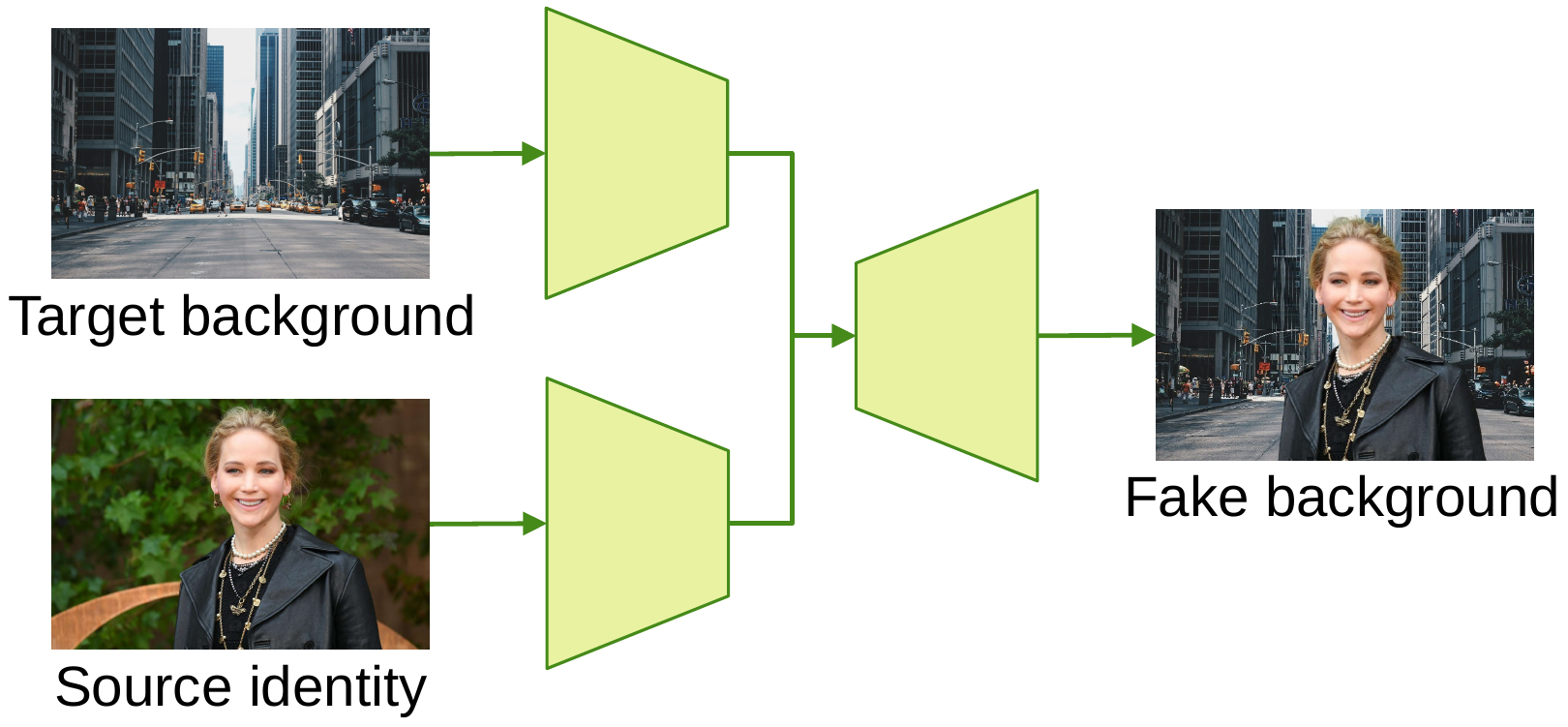}}
        \hfill
        \subfloat[Text-to-speech synthesis.\vspace{-0.15cm}]{\includegraphics[width=0.345\linewidth]{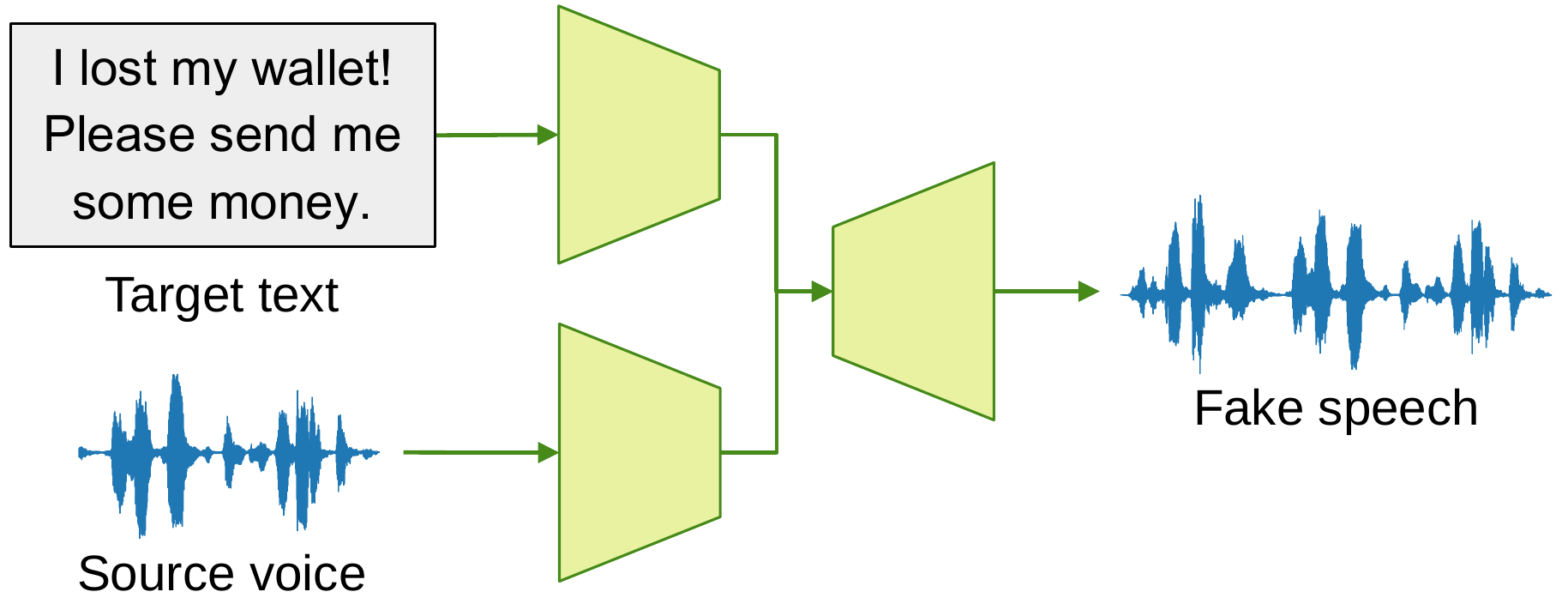}}
        \\
        \subfloat[Text-to-image generation.]{\includegraphics[width=0.31\linewidth]{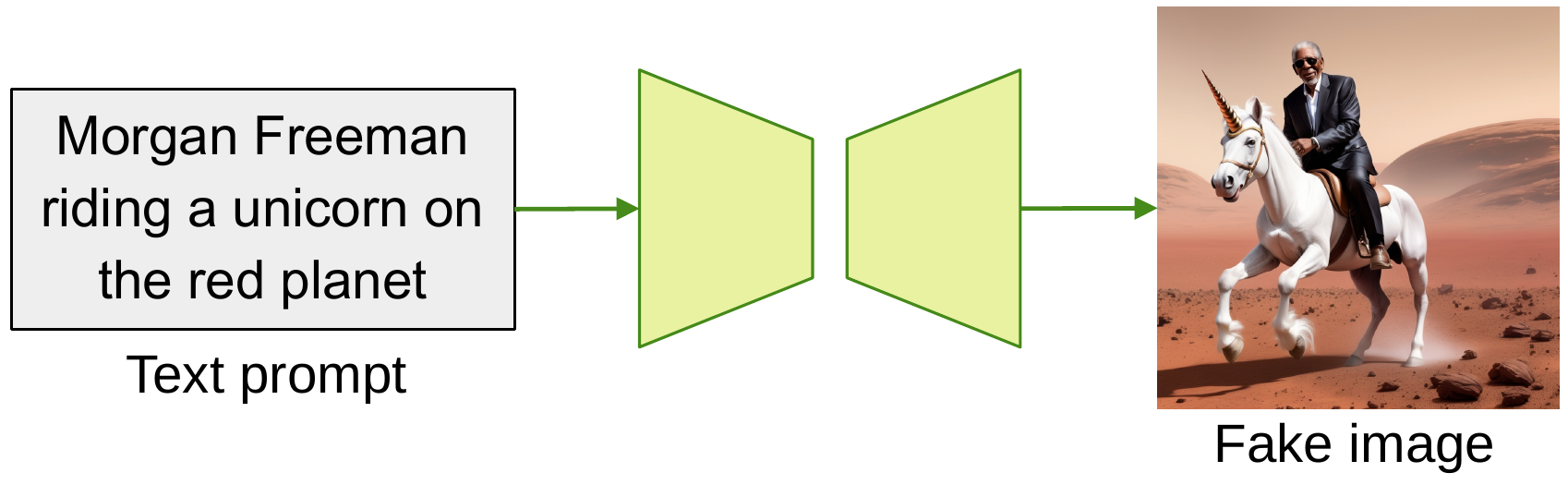}}
        \hfill
        \subfloat[Text-to-video generation.]{\includegraphics[width=0.31\linewidth]{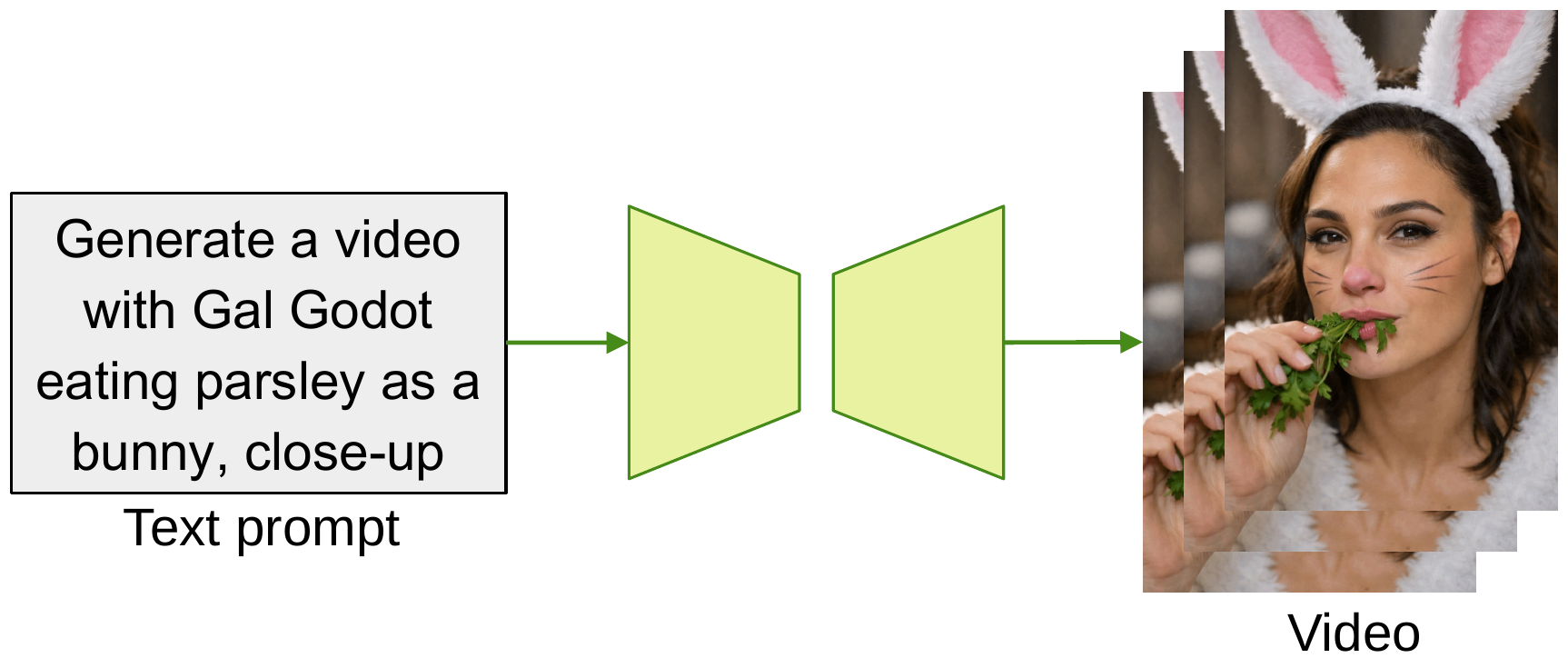}}
        \hfill
        \subfloat[Partial synthesis.]{\includegraphics[width=0.33\linewidth]{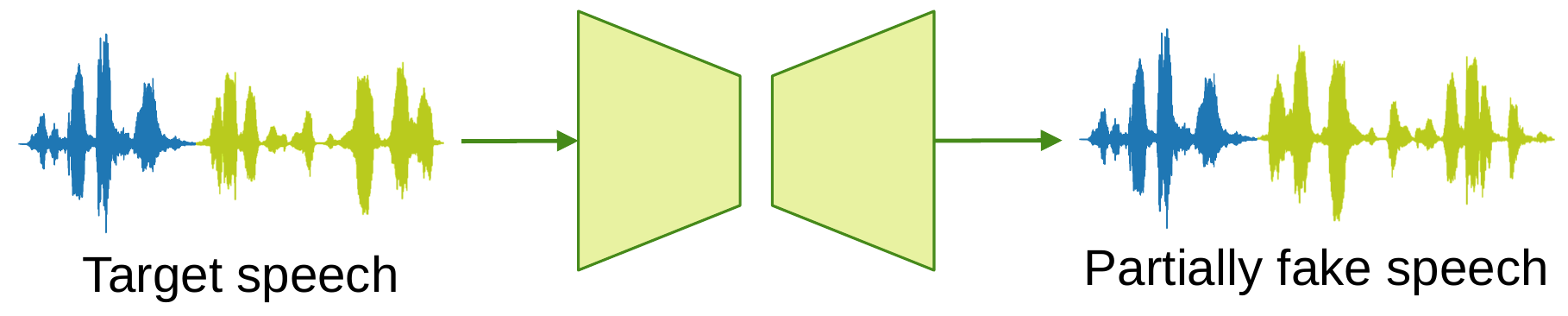}}
        \vspace{-0.3cm}
    \caption{Deepfake types according to the general procedure used to synthesize the fake content. For deepfake types that apply to multiple domains, we provide the illustration for only one domain. Best viewed in color.}
    \Description{This figure illustrates deepfake types according to the general procedure used to synthesize the fake content.}
    \label{fig_deepfake_types}
    \vspace{-0.3cm}
\end{figure}

\section{Deepfake Types}
\label{sec_deepfake_types}

To date, a number of alternative procedures have been employed to generate deepfakes. In order to simplify the task of producing realistic deepfakes, one commonly used procedure is to only alter a certain aspect of a media file, \eg~modifying the identity of a person in an existing video, or changing the emotion of a speech, while preserving the speech content and the speaker's identity. 
By employing recent and powerful generative models \cite{Rombach-CVPR-2022,Brooks-TR-2024}, generating deepfake content from scratch has also become prevalent. We further categorize the deepfake content according to the procedure employed to obtain the respective deepfake type. We illustrate the identified categories in Figure \ref{fig_deepfake_types} and present them in detail below. Interestingly, we identify deepfake categories that are domain-agnostic (which have been applied to all media types) and domain-specific (which have only been applied to a certain media type).

\paragraph{Identity swapping.} Deepfakes based on identity swapping imply replacing the identity information of a target person with that of a source person~\cite{Bitouk-SIGGRAPH-2008, Korshunova-ICCV-2017, Wang-IJCAI-2021, Chen-ACMMM-2020}, while preserving identity-agnostic attributes, such as facial expressions. In the visual domain, this kind of deepfake is often referred to as \emph{face swapping}, while in the audio domain, it is known as \emph{voice conversion} or \emph{voice swapping}. Voice conversion seeks to change the timbre and prosody of a speaker with those of another speaker, while preserving the content of the speech. 

\paragraph{Expression/emotion swapping.} In contrast to identity swapping, expression or emotion swapping~\cite{nirkin-ICCV-2019, Bounareli-ICCV-2023, Liu-CVPR-2024, rochow-CVPR-2024} involves altering the facial expression/emotion without changing the identity information. In the image domain, this task is known as \emph{facial expression swapping}. In the video domain, the task is also known as \emph{face reenactment}, and it implies altering the facial movement, which might often require facial motion capturing technology. In the audio domain, \emph{emotion swapping} is the task of changing the emotion of speech, while retaining the speech content and the speaker's identity.

\paragraph{Facial attribute manipulation.} Producing deepfake image and video content via facial attribute manipulation~\cite{Zhu-CVPR-2017, Choi-CVPR-2018, richardson-CVPR-2021} implies changing certain semantic attributes of a target face, while maintaining the identity information. Some of the attributes that are usually altered are age, gender, skin color and hair.

\paragraph{Talking face synthesis.} Talking face synthesis~\cite{gao-CVPR-2023b, Wang-CVPR-2023b, Zhong-CVPR-2023, Cho-MM-2024} is perhaps the most complex procedure to obtain deepfake audio-video content, but also the most flexible procedure. The task seeks to generate an audio-video file of a talking face, where the source character is engaged in the act of speech. The generated content is conditioned on some target information, provided in the form of text, audio, video or even multimodal content. The facial expressions, head movements, lip movements, speech emotions, and spoken content in the synthesized audio-video content are consistent with those of the source character.

\paragraph{Background swapping.} Deepfakes based on background swapping~\cite{Wang-CVPR-2024, Wang-CVPR-2024} are generated by changing the background scene with a different one. In the visual domain, this involves segmenting the source person and blending this person in a new scene. In the audio domain, the background sound of the original recording is replaced with another background sound using audio editing technologies, while preserving speech content and the speaker's identity.

\paragraph{Text-to-speech synthesis.} Deepfake audio can be created with the help of a machine learning model for speech synthesis, starting from a piece of text. Current text-to-speech (TTS) synthesis models~\cite{Li-NeurIPS-2023, Zen-Interspeech-2019, Kim-ICML-2021} can produce natural speech and emulate the voice of a source identity.

\paragraph{Text-to-image/video generation.} With the recent development of diffusion models~\cite{Rombach-CVPR-2022, singer-ICLR-2022, Nichol-ICML-2022, Chen-ICLR-2024}, such as Stable Diffusion \cite{Rombach-CVPR-2022} and GLIDE \cite{Nichol-ICML-2022}, a new type of deepfake has emerged. Deepfake content can be easily generated by simply prompting a text-conditional diffusion model. All the necessary details, including the name of the source person, the facial attributes, the body pose and the actions can be specified through the prompt. This approach can be used to generate both images and videos. In text-to-speech synthesis, the input text is literally pronounced by the system, while in text-to-image/video generation, the prompt is rather interpreted by the system as a set of instructions.

\paragraph{Partial synthesis.} As the name implies, partial synthesis~\cite{Rombach-CVPR-2022, Nichol-ICML-2022, Cheng-SIGGRAPH-2022} involves changing only a part of an existing media file to create a deepfake. In the video domain, this kind of deepfake can be obtained by changing a subset of frames. In the audio domain, partial synthesis seeks to change only a subset of words in an utterance. The changes are performed in such a way that the target identity is maintained for the entire duration of the video or audio clip.

\section{Deepfake Generation}

We organize our literature review of deepfake generation methods according to taxonomy illustrated in Figure A1 (in supplementary). We find that the most prominent approaches for deepfake generation are based on GANs or diffusion models. In some domains, such as video and audio, transformer-based methods are also very popular. Less frequently encountered methods are based on Variational Autoencoders (VAEs), Neural Radiance Fields (NeRF), 3D Morphable Models (3DMMs) and CNNs. Some models are only applied to specific media types, \eg~NeRF and 3DMMs are typically applied in the video domain. A number of studies use hybrid models, combining GANs with VAEs on the one hand, or CNNS with RNNs on the other. We start by providing tutorials for the main generative frameworks below. 
We structure the remainder of our presentation according to the media type. For each media type, we divide the surveyed studies according to the underlying architectures. Finally, we compare deepfake generations methods in terms of visual quality, performance and speed. 

\subsection{Deepfake generation tutorial}\label{sec:supp_introduction}

Various deep generative models are actively being used to successfully generate deepfake content. Among the deepfake generative methods, we next explain in detail how the most popular and interesting developments work, such as generative adversarial networks, variational autoencoders, diffusion models, as well as Neural Radiance Fields. We provide usage examples for each generative framework in Section \ref{sec_supp_A} from the supplementary material.

\subsubsection{Generative Adversarial Networks}

Generative adversarial networks (GANs)~\cite{Goodfellow-NIPS-2014} consist of two neural networks, called the generator and the discriminator. The generator, denoted as $G_{\theta}(z)$, transforms input Gaussian noise $z \sim p(z)$ into a sample from the data distribution. The discriminator, represented as $D_{\phi}(x)$, outputs a single scalar value that predicts the probability of a given sample $x$ to be real, rather than being generated by $G_{\theta}$. Therefore, the discriminator $D_{\phi}$ is trained as a binary classifier, where the real samples are labeled with $1$ and the generated ones with $0$, as follows:
\begin{equation}
\begin{split}
     \min_{D_\phi}\mathcal{L}_{D}=&-\mathbb{E}_{x\sim p(x)}\!\left[\log \left(D_\phi(x)\right)\right] -\mathbb{E}_{z \sim p(z)}\!\left[\log\left(1\!-\!D_\phi(G_\theta(z))\right)\right],
\end{split}
\end{equation}
where $p(x)$ and $p(z)$  represent the real data distribution and the Gaussian distribution, respectively, with $p(z)$ serving as a source for sampling varied inputs for the generator.
The generator is trained to deceive the discriminator. Thus, its training objective maximizes the probability assigned by the discriminator to the generated samples $D_{\phi}(G_{\theta}(z))$:
\begin{equation}
    \min_{G_\theta}{\mathcal{L}_{G}=-\mathbb{E}_{z \sim p(z)}\left[ \log\left(D_\phi(G_\theta(z))\right)\right]}.
\end{equation}
We can also express the whole training framework through a single mini-max optimization objective, as follows:
\begin{equation}
\begin{split}
    \min_{G_\theta}\max_{D_\phi}&\mathbb{E}_{x\sim p(x)}\!\left[\log\!\left(D_\phi(x)\!\right)\right]\!
    +\mathbb{E}_{z \sim p(z)}\!\left[\log\!\left(1\!\!-\!\!D_\phi(G_\theta(z)\!)\!\right)\right],
\end{split}
\end{equation}
Note that optimizing for $G_{\theta}$ does not influence the first term of the objective, as it depends only on $D_\phi$.

\subsubsection{Variational Autoencoders}

A Variational Autoencoder (VAE)~\cite{Kingma-arxiv-2013} is a modified version of the classic autoencoder, designed to support generative modeling by learning a probabilistic latent space. In the classic autoencoder case, during training, an encoder maps a sample $x$ to a latent representation $z$ from which a decoder is tasked to reconstruct the original input $x$. This approach is unsuitable for generative modeling because $z$ follows an arbitrary complex distribution, making it impossible to directly sample from it. Kingma~\etal~\cite{Kingma-arxiv-2013} address this limitation by enforcing $z$ to follow a standard Gaussian distribution. They achieved this by adding a Kullback-Leibler (KL) divergence regularization term to the loss function, which minimizes the divergence between the distribution of $z$ and the standard Gaussian distribution. To model the distribution of $z$, the encoder outputs a mean $\mu_x$ and a variance $\sigma_x$ that describe a Gaussian distribution. During training $z$ is sampled from this distribution. The loss function is as follows:
\begin{equation} 
\begin{split}
        \mathcal{L} = &\mathbb{E}_{z\sim E_\phi(z|x)}\left[\lVert D_\theta(z) - x\rVert_2^2\right] + \mbox{KL}\left(E_\phi(z|x)\lVert p(z)\right),
\end{split}
\end{equation}
where $E_\phi(z|x)$ denotes the encoder, $D_\theta(x|z)$ is the decoder and $p(z)$ represents the standard Gaussian distribution.

\subsubsection{Diffusion models}

Diffusion models~\cite{Croitoru-TPAMI-2023} consist of two diffusion processes, the forward diffusion process and the reverse diffusion process. In the forward process, a data sample is progressively transformed over 
$T$ steps by adding Gaussian noise at each step, eventually converting it into an approximate standard Gaussian distribution. The reverse process operates in the opposite direction, serving as the generative mechanism. It learns to map a standard Gaussian sample back to a data sample.

\paragraph{Forward process.} The forward process is a Markov chain $x_t \sim q(x_t|x_{t-1}) = \mathcal{N}(\sqrt{1-\beta_t}\cdot x_{t-1}, \beta_t \cdot \mathbf{I})$, which gradually adds Gaussian noise that depends on a variance schedule $\{\beta_t\}_{t=1}^{T}$, where $x_0 \sim q(x_0)$ represents a data sample and $x_T$ is approximately a standard Gaussian sample. This formulation supports an efficient sampling for an arbitrary $x_t$ during training:
\begin{equation}
    x_t \sim \mathcal{N}(\sqrt{\bar{\alpha}_t}x_0, (1-\bar{\alpha}_t)\cdot \mathbf{I}),
\end{equation}
where $\alpha_t = 1-\beta_t$, $\bar{\alpha}_t=\prod_{i=0}^{t}\alpha_i$.

\paragraph{Reverse process.} The reverse process starts with $x_T\sim\mathcal{N}(0, \mathbf{I})$ and follows the learned Gaussian transitions denoted as $p_\theta(x_{t-1}|x_t) = \mathcal{N}(\mu_\theta(x_t, t), \Sigma_\theta(x_t, t))$ to recover a data sample $x_0$. Commonly, in practice, the variance $\Sigma_\theta(x_t, t)$ is approximated with $\beta_t$. Thus, the only learnable component that remains is the mean $\mu_\theta(x_t, t)$, which can be rewritten as a function of noise~\cite{Ho-NeurIPS-2020}:
\begin{equation}
    \mu_\theta(x_t, t) = \frac{1}{\sqrt{\alpha_t}}\left(x_t - \frac{\beta_t}{\sqrt{1-\bar{\alpha_t}}}\cdot\epsilon_\theta(x_t, t)\right).
\end{equation}
As a consequence, during training, the neural network, denoted by $\epsilon_\theta(x_t, t)$, learns to approximate the noise $\epsilon \sim \mathcal{N}(0, \mathbf{I})$ added at arbitrary steps $t\sim \mathcal{U}(1,\dots, T)$ to the data samples $x_0 \sim q(x_0)$. Formally, the optimization objective for the reverse process is defined as:
\begin{equation} 
    \min_{\epsilon_\theta}\mathcal{L}_{simple} = \mathbb{E}_{x_0, t, \epsilon }  \lVert \epsilon - \epsilon_\theta(x_t, t) \rVert^2_2.
\end{equation}
where $x_0 \sim q(x_0)$, $t\sim \mathcal{U}(1,\dots, T)$ and $\epsilon \sim \mathcal{N}(0, \mathbf{I})$.

\subsubsection{Neural Radiance Fields}

Neural Radiance Fields (NeRFs)~\cite{Mildenhall-CACM-2021} represent a method for synthesizing novel views from a sparse set of input images of a scene. The core idea is to train a single neural network to overfit to a specific scene, with the weights encoding the detailed information and structure of the scene. The input to the neural network is a 5D vector consisting of three spatial coordinates $(x, y, z)$, which represent a point in 3D space of the scene, and two angles $(\theta, \phi)$, which define the viewing direction. The network outputs the density $\sigma$ at the given spatial point and its color $c$, represented as an RGB vector. However, datasets typically lack direct annotations for the density and color of each spatial point, so the network optimization is performed in the pixel space instead. This process involves selecting a viewing direction and casting a ray through the scene, along which multiple spatial points are sampled. These points are processed by the neural network to compute their corresponding density and color values. The outputs are then combined using volume rendering to generate an approximate pixel value. This predicted pixel value is compared with the ground-truth pixel using a mean squared error loss function to optimize the network. Formally, the loss function is defined as:
\begin{equation}
    \mathcal{L} = \Sigma_{r \sim \mathcal{R}}\lVert\hat{C}(r) - C(r)\rVert_2^2,
\end{equation}
where $\mathcal{R}$ is the set of rays in the current batch, $\hat{C}(r)$ is the predicted color for a particular ray $r$ and $C(r)$ is the ground-truth color.

In practice, NeRFs have two implementation details for a better optimization. First, spatial positions are projected into a higher-dimensional space using a positional encoding, similar to the approach used for vision transformers. Second, NeRFs use Hierarchical Volume Sampling, which involves training two neural networks. The first network samples coarse spatial points along a ray to estimate the overall structure of the scene. It then identifies regions of greater importance. The second network focuses on these regions, performing finer sampling to achieve more accurate results.

\subsection{Image} 

\subsubsection{GAN-based methods} 

\paragraph{Face synthesis.}
Creating realistic faces is essential for deepfake generation, and GANs are widely employed to achieve this \cite{Shen-CVPR-2018, karras-ICLR-2017, Karras-TPAMI-2021, Karras-CVPR-2020, Karras-NeurIPS-2021, Esser-CVPR-2021, Sauer-SIGGRAPH-2022, Fu-ECCV-2022, Xu-CVPR-2023}. Shen \etal~\cite{Shen-CVPR-2018} introduce a third model into the traditional adversarial framework, tasked with determining whether the generated images retain the identity from a reference image. This approach enables the method to perform conditional generation. Conditional generation is also the focus of Xu \etal~\cite{Xu-CVPR-2023}. Their approach synthesizes high-quality 3D heads with control over the camera poses and other facial attributes. StyleGAN~\cite{Karras-CVPR-2020, Karras-TPAMI-2021, Karras-NeurIPS-2021} improves the quality of the synthesized images by changing the generator architecture, and leveraging a mapping network to map the usual Gaussian vector to an intermediary latent space. 
Fu \etal~\cite{Fu-ECCV-2022} demonstrate that StyleGAN is also effective for generating full body images. Sauer \etal~\cite{Sauer-SIGGRAPH-2022} extend the StyleGAN model, presenting a method that leverages Projected GAN training~\cite{Sauer-NeurIPS-2021}, progressive growing and classifier guidance~\cite{Dhariwal-NeurIPS-2021}, unlocking image synthesis at a resolution of 1024$\times$1024. 

\paragraph{Face swapping.} One of the most widely-used methods for generating deepfakes is face swapping. This technique involves replacing the face in a target image with that of another individual, sourced from a different image. The key challenge lies in seamlessly integrating the source face, while maintaining non-identity-specific attributes, such as facial expressions and lighting conditions. Thanks to their well-known capacity of generating realistic images, GANs~\cite{Goodfellow-NIPS-2014, karras-ICLR-2017, Bao-CVPR-2018, Chen-ACMMM-2020, Li-CVPR-2023, Natsume-SIGGRAPH-2018, Ren-ICCV-2023, Liu-CVPR-2023, Kim-CVPR-2022, Shiohara-CVPR-2023, Rosberg-WACV-2023, Li-CVPR-2020} are widely adopted in face swapping frameworks.

In GAN-based face swapping pipelines, the generator is usually conditioned on identity information from the source image and the attributes extracted from the target image \cite{Bao-CVPR-2018, Li-CVPR-2020, Chen-ACMMM-2020, Kim-CVPR-2022, Shiohara-CVPR-2023, Rosberg-WACV-2023, Zeng-AAAI-2023, Yoo-WACV-2023, Cui-CVPRW-2023, Yuan-ArXiv-2023}. The work of Bao \etal~\cite{Bao-CVPR-2018} is one of the earliest contributions in this direction. The authors employ a face recognition model to extract an identity embedding from the source image. The attributes of the target image are extracted by a neural network trained to minimize the Euclidean distance between the target and generated images, applying a lower weight when the identities in the target and source images differ. 
Li \etal~\cite{Li-CVPR-2020} advance the previous framework by introducing a multi-level attribute encoder that is trained in a self-supervised manner, providing a more detailed representation of the target image than the approach of Bao \etal~\cite{Bao-CVPR-2018}. 
Similarly, Chen \etal~\cite{Chen-ACMMM-2020} propose the ID Injection Module, which integrates identity information through Adaptive Instance Normalization (AdaIN) \cite{Huang-ICCV-2017} layers into the target image features. More recent works \cite{Kim-CVPR-2022, Shiohara-CVPR-2023, Rosberg-WACV-2023, Cao-FG-2023, Zeng-AAAI-2023, Yoo-WACV-2023} increase the quality and quantity of the conditional identity information. Kim \etal~\cite{Kim-CVPR-2022} enforce smoothness to the identity encoding space through contrastive learning, while Rosberg \etal~\cite{Rosberg-WACV-2023} leverage the feature maps provided by multiple layers of the face encoder to better represent the identity. Cao \etal~\cite{Cao-FG-2023} and Cui \etal~\cite{Cui-CVPRW-2023} take a different route by exploring the effectiveness of the transformer architecture for identity embedding in face swapping. 
Zeng \etal~\cite{Zeng-AAAI-2023} go a step further, showing that a masked autoencoder (MAE)~\cite{He-CVPR-2022}, pre-trained on a large-scale face dataset, is an effective encoder for face swapping. 

Slightly distinct from earlier studies, another line of research \cite{Natsume-SIGGRAPH-2018, Gao-CVPR-2021, Zhu-CVPR-2021, Li-CVPR-2023, Jiang-CVPR-2023, Ren-ICCV-2023, Liu-CVPR-2023} explores the manipulation of identity and attribute features within the latent space of the generator. Natsume \etal~\cite{Natsume-SIGGRAPH-2018} create the latent space of the generator by merging the outputs of two encoders. One of the encoders is responsible for identity information, and the other for attributes. 
Two encoders are also utilized by Ren \etal~\cite{Ren-ICCV-2023} to separately learn embeddings for facial non-identity and non-facial attributes. This separation eliminates the need for skip connections, preventing identity leakage from the target image. 
Zhu \etal~\cite{Zhu-CVPR-2021} train a model to perform GAN inversion and obtain the latent code for a given image. Subsequently, they employ another model to integrate the attributes from the target image into the latent code of the source face. The resulting latent code is fed into StyleGAN2~\cite{Karras-CVPR-2020} to generate the swapped image. Similarly, Li \etal~\cite{Li-CVPR-2023} employ learnable GAN inversion, but in their case, they leverage the latent space of a 3D GAN~\cite{Chan-CVPR-2021} to synthesize multi-view swapped images. The latent space of StyleGAN is further exploited for face swapping by Liu \etal~\cite{Liu-CVPR-2023}, where the GAN inversion is extended at region level through the use of facial semantic masks. 

GANs are also used in face swapping for the purpose of fixing the swapped image and making it more realistic~\cite{Moniz-NeurIPS-2018, Sun-ECCV-2018, Chen-CVPR-2021}. Specifically, Moniz \etal~\cite{Moniz-NeurIPS-2018} and Sun \etal~\cite{Sun-ECCV-2018} employ GANs to perform the blending of the source face in the target image, leveraging inpainting pipelines or frameworks such as CycleGAN~\cite{Zhu-CVPR-2017}. Chen \etal~\cite{Chen-CVPR-2021} take a step further and present a method to correct deepfake images perturbed with adversarial attacks.

\paragraph{Face editing.}
Altering facial attributes such as age, gender, hair color or pose can be used to create counterfeit content. GANs support this kind of applications, either via image-to-image translation between different domains~\cite{Zhu-CVPR-2017, Choi-CVPR-2018, Choi-CVPR-2020, Hsu-CVPR-2022} or via latent code manipulation~\cite{Tov-TOG-2021, Bounareli-ICCV-2023, Suwala-WACV-2024}. CycleGAN~\cite{Zhu-CVPR-2017} was first proposed for image-to-image translation between two domains. The method comprises two generators, one for each of the two domains. The primary contribution of Zhu \etal~\cite{Zhu-CVPR-2017} is the introduction of the cycle-consistency loss, which ensures that the pipeline can reconstruct the original image, after translating it from one domain to the other and back. The main limitation of CycleGAN is its ability to handle only two domains. StarGAN~\cite{Choi-CVPR-2018, Choi-CVPR-2020} addresses this limitation and supports image translation from multiple domains 
by including an additional condition as input, along with the conditional image. 
Hsu et al. \cite{Hsu-CVPR-2022} employ a dual-generator approach for image-to-image translation. The first generator produces a landmark image matching the pose of a reference image, and the second uses this landmark to recreate a source identity in the specified pose. Different from these approaches, other works~\cite{Tov-TOG-2021, Bounareli-ICCV-2023, Suwala-WACV-2024} harness the latent space of StyleGAN. For instance, Tov~\etal~\cite{Tov-TOG-2021} study the latent space of StyleGAN and design an encoder for inversion, which is suitable for image editing. Similarly, Suwa{\l}a \etal~\cite{Suwala-WACV-2024} design a plugin for the latent space of StyleGAN. This plugin disentangles the latent codes into attribute and non-attribute features, allowing attribute manipulation for facial editing. 

\subsubsection{Diffusion-based methods}

\paragraph{Text-to-image.} Diffusion models are effectively applied in text-to-image generation~\cite{Chen-ICLR-2024, Podell-ICLR-2024, Saharia-NeurIPS-2022, Rombach-CVPR-2022}, utilizing large language models to encode textual descriptions that guide image creation. This capability allows users to generate counterfeit content featuring public figures simply by including their names in the text description used as input for generation. One of the most popular methods for text-to-image generation is Stable Diffusion~\cite{Rombach-CVPR-2022}, which leverages the latent space of a vector quantized (VQ) GAN~\cite{Esser-CVPR-2021} to perform the diffusion processes. SDXL~\cite{Podell-ICLR-2024} scales up the Stable Diffusion architecture, improving the quality and text fidelity of the generated images.

\paragraph{Personalized generation.} Although text-to-image diffusion models allow deepfake content generation of public figures, some results do not accurately replicate the identity of the person. Thus, these models might have limited application in deepfake generation. However, there is another direction of research~\cite{Zhao-CVPR-2023, Ruiz-CVPR-2023, Chen-AAAI-2024, Peng-CVPR-2024b, Ma-SIGGRAPH-2024, Liu-ArXiv-2023, Boutros-ICCV-2023, Han-ArXiv-2023, Lin-CVPR-2024a, Wang-ArXiv-2024b, Wang-CVPR-2024, Wu-ArXiv-2024, Kim-ArXiv-2024, Wang-ArXiv-2024c, Liu-CVPR-2024, Guo-NeurIPS-2024, Gu-ECCV-2024, Wang-ArXiv-2024d, Papantoniou-ECCV-2024, Wang-ArXiv-2024e, Chen-ArXiv-2024b, Huang-ArXiv-2024b, Huang-ArXiv-2024a, He-ArXiv-2024, Li-CVPR-2024a, Wei-ECCV-2024} focused on generating images that contain a specific identity or concept depicted in an image or a set of images given as input. Such methods are more likely to be employed in deepfake generation. We can distinguish these contributions into two main approaches, those that perform test-time fine-tuning~\cite{Liu-ArXiv-2023, Lin-CVPR-2024a, Wang-ArXiv-2024c, Gu-ECCV-2024, Gal-ICLR-2023, Ruiz-CVPR-2024, Chen-NeurIPS-2023, Ruiz-CVPR-2023} and those that leverage large-scale datasets and learn how to incorporate the additional images offline~\cite{Zhao-CVPR-2023, Han-ArXiv-2023, Chen-AAAI-2024, Peng-CVPR-2024b, Ma-SIGGRAPH-2024, Boutros-ICCV-2023, Huang-ArXiv-2024a, Wang-ArXiv-2024b, Wang-CVPR-2024, Wu-ArXiv-2024, Kim-ArXiv-2024, Liu-CVPR-2024, Gu-ECCV-2024, Wang-ArXiv-2024d, Papantoniou-ECCV-2024, Wang-ArXiv-2024e, Chen-ArXiv-2024b, Huang-ArXiv-2024b, He-ArXiv-2024, Li-CVPR-2024a,  Wei-ECCV-2024}.

Test-time fine-tuning approaches use different components to integrate and learn the new identity. Some works introduce a new text token for the identity and learn to embed it~\cite{Lin-CVPR-2024a, Wang-ArXiv-2024c, Gal-ICLR-2023}. Other approaches~\cite{Liu-ArXiv-2023} either use low-rank adaptation (LoRA)~\cite{Hu-ICLR-2022} or directly fine-tune the weights of the denoising network~\cite{Gu-ECCV-2024, Ruiz-CVPR-2024, Chen-NeurIPS-2023, Ruiz-CVPR-2023}. Overall, test-time fine-tuning methods yield impressive results in terms of identity preservation, but their main disadvantage is the expensive optimization, which significantly increases the generation time. To this end, many works address the efficiency issue, to some extent. For example, Chen~\etal~\cite{Chen-NeurIPS-2023} try to incorporate the knowledge of multiple subject-specific models into a single model. Ruiz~\etal~\cite{Ruiz-CVPR-2024} leverage a HyperNetwork architecture to predict the network weights from a face image. Subsequently, they use these weights as a starting point for test-time fine-tuning, reaching faster convergence than previous work~\cite{Ruiz-CVPR-2023}. Despite these advancements, test-time fine-tuning methods still suffer from high generation times.

In contrast to test-time fine-tuning methods, the approaches that harness offline training~\cite{Zhao-CVPR-2023, Chen-AAAI-2024, Peng-CVPR-2024b, Ma-SIGGRAPH-2024, Boutros-ICCV-2023, Han-ArXiv-2023, Huang-ArXiv-2024a, Wang-ArXiv-2024b, Wang-CVPR-2024, Wu-ArXiv-2024, Kim-ArXiv-2024, Liu-CVPR-2024, Gu-ECCV-2024, Wang-ArXiv-2024d, Papantoniou-ECCV-2024, Wang-ArXiv-2024e, Chen-ArXiv-2024b, Huang-ArXiv-2024b, He-ArXiv-2024, Li-CVPR-2024a} are faster in terms of generation time, but their primary issue is identity preservation. Therefore, solving the latter problem constitutes the priority of these works. Zhao~\etal~\cite{Zhao-CVPR-2023} propose an identity preservation loss for which they construct a better estimation of the original image given the predicted noise, at training time. The same idea is studied by Liu~\etal~\cite{Liu-CVPR-2024}, who improve the estimation even further. Peng~\etal~\cite{Peng-CVPR-2024b} employ an identity loss, but only for certain noise levels. Other methods~\cite{Papantoniou-ECCV-2024}, inspired by the GAN literature, employ the ArcFace model as identity embedding extractor for better identity representations. Similarly, Li~\etal~\cite{Li-CVPR-2024a} improve representations by stacking multiple embeddings when multiple images are available. Lastly, Wang~\etal~\cite{Wang-CVPR-2024} decouple the generation of background and identity-related content by training two separate denoising networks, out of which only one knows how to generate images of a given person. 

\paragraph{Tools.} The most popular tools for personalized generation are LoRA-based variants of Stable Diffusion~\cite{Rombach-CVPR-2022} and SDXL~\cite{Podell-ICLR-2024}, that are specialized on particular public personalities, \eg~Elon Musk\footnote{\href{https://civitai.com/models/603798/elon-musk-sdxl}{https://civitai.com/models/603798/elon-musk-sdxl}} or Alan Turing\footnote{\href{https://civitai.com/models/796450/alan-turing-mathematical-flux}{https://civitai.com/models/796450/alan-turing-mathematical-flux}}. Different from these options, another powerful tool is Midjourney\footnote{\href{https://www.midjourney.com/home}{https://www.midjourney.com/}}. In contrast to the LoRA-based methods, Midjourney is not popular for personalized generation, but for text-to-image synthesis. However, given the quality of its generative results, Midjourney is a popular tool for creating counterfeit images.

\subsubsection{Other methods}

Unlike previous methods centered around generative models, some approaches rely on alternative techniques.
Bitouk \etal~\cite{Bitouk-SIGGRAPH-2008} identify the closest match in terms of lighting and pose from a large set of face images, and perform face replacement using key point alignment.
Korshunova \etal~\cite{Korshunova-ICCV-2017} use a multi-resolution CNN in the VGG feature space, aligning target and source images to minimize cosine distance between corresponding patches. Wang \etal~\cite{Wang-IJCAI-2021} propose an encoder-decoder architecture with a 3D identity extractor and a Semantic Facial Fusion module to enhance resolution and preserve identity.
In contrast, other works combine generative methods to produce higher-quality images. Bao \etal~\cite{Bao-ICCV-2017} introduce CVAE-GAN, a method which combines VAEs with GANs. 
The generator and the encoder are trained with an adversarial objective, but also with a mean feature matching objective and a pixel-wise reconstruction loss, respectively. Li \etal~\cite{Li-CVPR-2024b} merge diffusion models and GANs by representing identity in Stable Diffusion through the latent space of StyleGAN, integrating latent codes into the U-Net via cross-attention layers.





\subsection{Video}

\subsubsection{GAN-based methods}

The early works for generating deepfake videos employ conventional GAN models for face swapping and reenactment, which are applied frame by frame \cite{nirkin-ICCV-2019, xu-ECCV-2022}. Nevertheless, these methods are usually part of more complex frameworks which have zero-shot capabilities, either involving more steps 
\cite{nirkin-ICCV-2019} or enhanced architectures \cite{xu-ECCV-2022}. Gao \etal~\cite{gao-CVPR-2023b} introduce a face reenactment GAN, focusing on generating videos of talking heads. The facial landmarks, expressions and head poses are extracted from both source and target frames to fit a face 3DMM and obtain predefined keypoints. 

To depart from the conventional paradigm and improve the video generation using GANs, subsequent works \cite{brooks-NeurIPS-2022,skorokhodov-CVPR-2022, tian-ICLR-2021, tulyakov-CVPR-2018, oorloff-ICCV-2023, yu-ICLR-2022} leverage the latent space of StyleGAN2 \cite{Karras-CVPR-2020}. A consistent number of methods divide the latent space in which they operate into two: one for content and one for motion \cite{skorokhodov-CVPR-2022, tian-ICLR-2021, tulyakov-CVPR-2018}. While some utilize an RNN for sampling the motion trajectory \cite{tian-ICLR-2021, tulyakov-CVPR-2018} and employ two discriminators, one for individual frames and one for the video sequence, Skorokhodov \etal~\cite{skorokhodov-CVPR-2022} compute the motion embeddings with 1D convolutional layers and use only one video discriminator. Oorloff \etal~\cite{oorloff-ICCV-2023} take a different approach by encoding both source and target frames, fusing their latent representations, then generating the output frame, while also utilizing multiple latent spaces \cite{abdal-ICCV-2019,wu-CVPR-2021}. 

\subsubsection{Diffusion-based methods}

Latent diffusion models \cite{Rombach-CVPR-2022} use a cross-attention mechanism that facilitates conditioning diffusion models for image generation. However, the main challenge in generating deepfake videos with diffusion models is employing a conditioning mechanism, while achieving temporal cohesion.
The studies of Ho \etal~\cite{ho-ArXiv-2022b} and Blattman \etal~\cite{blattmann-CVPR-2023} represent the stepping stones in adopting diffusion models for video generation. Their methods extend diffusion models in several ways. The architectural changes applied on the U-Net mainly consist of replacing 2D convolutions with 3D convolutions, and appending additional self-attention layers for temporal attention. In a subsequent work, Blattmann \etal~\cite{blattmann-ArXiv-2023} demonstrate the benefits of using a large curated dataset for training a video generator. 
Wu \etal~\cite{wu-ICCV-2023} introduce a one-shot method for editing a video given a text prompt. Inspired by Ho \etal~\cite{ho-ArXiv-2022b}, a text-to-image diffusion model is extended to an additional dimension (time) by Wu \etal~\cite{wu-ICCV-2023}, where the added self-attention layers operate on the current frame and the previous two frames. 

Newer diffusion-based video generation methods \cite{bao-ArXiv-2024, guo-ICLR-2024, ma-ArXiv-2024a, Xing-CVPR-2024, Xu-MM-2025, Ding-ICLR-2026} depart from the U-Net architecture and adopt a transformer-based one, namely ViT, which provides an innate mechanism for both spatial and temporal attention. This allows longer videos to be generated. Additionally, Guo \etal~\cite{guo-ICLR-2024} introduce a plug-and-play module that can be integrated into a text-to-image diffusion model to induce the ability to generate videos. Inspired by this module, Wang \etal~\cite{wang-ArXiv-2024} present a method for text-to-video generation composed of several stages, in which a ControlNet is applied to improve guidance.
Different from previous studies employing Stable Diffusion as the base model, Singer \etal~\cite{singer-ICLR-2022} ground their work on DALLE-2 \cite{ramesh-ArXiv-2022}, while Ho \etal~\cite{ho-ArXiv-2022a} utilize Imagen \cite{Saharia-NeurIPS-2022}. Nevertheless, similar architectural changes are implemented, where the network is extended to support the temporal dimension.

An important line of research is represented by portrait animation, in which a video is generated from a source frame and various conditional inputs. Most works in this area \cite{bounareli-ArXiv-2024, ma-ArXiv-2024b, yang-ArXiv-2024, zhang-ArXiv-2024, Ma-IJCV-2026, Lin-CVPR-2025, Lin-MM-2025} aim to apply a sequence of facial expressions over the image. Two different approaches are used to condition the diffusion model. One is based on intermediate representations of the facial expressions, such as facial keypoints \cite{bounareli-ArXiv-2024, ma-ArXiv-2024b, zhang-ArXiv-2024}, and the other is based on directly encoding the frames containing the target facial movements \cite{yang-ArXiv-2024}.

Currently, the ability of the video generation methods based on diffusion modeling is not satisfactory, often requiring quality enhancements at a later stage in the pipeline. For example, super-resolution models are sometimes employed to increase video resolution \cite{ho-ArXiv-2022a, singer-ICLR-2022, wang-ArXiv-2024}, while the frame rate is usually increased through frame interpolation \cite{blattmann-ArXiv-2023, ma-ArXiv-2024b, singer-ICLR-2022, wang-ArXiv-2024}.

\subsubsection{Other methods}

Transformers represent the most popular architectural choice for video generation. For instance, Rochow \etal~\cite{rochow-CVPR-2024} leverage cross-attention blocks to guide the generation (using encoded facial keypoints and expressions), while Villegas \etal~\cite{villegas-ICLR-2023} apply the attention mechanism on frames to generate longer and coherent videos.

An alternative choice for video generation consists of employing some VQ autoencoder, either variational \cite{van-NeurIPS-2017} or standard \cite{yu-CVPR-2023}. Similar to diffusion models, the generation process is carried out in the latent space of the autoencoder, which is lower dimensional. Within this vector space, a transformer is used to generate video tokens \cite{hong-ICLR-2023, jiang-ICCV-2023, yan-ArXiv-2021, yu-CVPR-2023}. Unlike other related approaches, Jiang \etal~\cite{jiang-ICCV-2023} carefully design the latent space such that it is decomposed into an appearance and a pose representation, respectively.

A few methods harness the 3D space for face reenactment. In this context, warping is often employed, which involves computing a flow field between the source frame and the driving frame, then applying it on the former frame. In the same context, NeRF models \cite{Mildenhall-CACM-2021} are used to generate novel views of 3D face models. For example, Thies \etal~\cite{Thies-TOG-2019} obtain a coarse 3D representation from the source frame using a traditional graphics pipeline, and then feed it to a neural network to obtain a neural texture, a high-dimensional embedding space, from which a Deferred Neural Renderer (based on U-Net) generates the target image. Thies \etal~\cite{thies-CVPR-2016} and Yang \etal~\cite{yang-ECCV-2022} synthesize faces by applying a deformation transfer between two 3DMM-based intermediate representations of the source and driving video frames, the mouth being further refined through warping. Zhang \etal~\cite{zhang-CVPR-2023} also apply warping based on dense landmarks, while Li \etal~\cite{li-CVPR-2023b} combine warping with NeRF. Finally, to increase the performance, some works adopt pre-training strategies that involve masking the input and then reconstruct the signal \cite{villegas-ICLR-2023, yu-CVPR-2023}.

\subsection{Audio}

\subsubsection{GAN/VAE-based methods}

A number of text-to-speech models employ popular generative frameworks \cite{Casanova-ICML-2022,Kim-ICML-2021,Tan-TPAMI-2024,Lee-ICLR-2023}, such GANs and VAEs, either alone or in combination with more recent developments in the field. Kim \etal~\cite{Kim-ICML-2021} propose an end-to-end TTS framework that augments variational inference with normalizing flows and uses an adversarial training procedure to enhance the representation potential. 
The method of Casanova \etal~\cite{Casanova-ICML-2022} constructs on the previous model, introducing new procedures, such as the concatenation of language embeddings with the ones of the input characters to allow training in a multilingual fashion. 

In \cite{Tan-TPAMI-2024}, the authors introduce new modules to develop NaturalSpeech, another VAE-based TTS. A differentiable durator is used to improve the duration prediction, a memory mechanism simplifies the waveform reconstruction, and a bidirectional prior/posterior module improves the prior from text, while simplifying the posterior from speech.
Lee~\etal~\cite{Lee-ICLR-2023} present a GAN-based vocoder that improves the generator by introducing anti-aliased feature representation and periodic non-linearities, delivering state-of-the-art results and robustness for out-of-distribution scenarios, such as novel languages and speakers.

\subsubsection{Transformer-based methods}

A few recent methods \cite{Jiang-ArXiv-2023,Kharitonov-TACL-2023,Ren-NeurIPS-2019,Wang-ArXiv-2023,Yang-ArXiv-2023} employ transformers to obtain competitive generation performance. FastSpeech \cite{Ren-NeurIPS-2019} introduces a transformer-based model that speeds up speech synthesis by parallelizing Mel-spectrogram generation through a feed-forward architecture. A length regulator is used to match the length of the hidden states with the length of the Mel-spectrograms, and a duration predictor provides the duration for the phonemes. 
Jiang \etal~\cite{Jiang-ArXiv-2023} reuse the length regulator and the duration predictor from FastSpeech, adding separate modules for content, timbre and prosody modeling. 

Wang \etal~\cite{Wang-ArXiv-2023} proposed VALL-E, a framework that uses intermediate representations consisting of audio codec codes instead of Mel-spectrograms. A pre-trained neural codec model generates acoustic codes that are used alongside corresponding phoneme sequences during training, allowing the neural language model to extract both speaker information and content. SPEAR-TTS \cite{Kharitonov-TACL-2023} removes the necessity to supply the transcripts of audio prompts by decoupling the generation of semantic tokens and the acoustic tokens. 


\subsubsection{Diffusion-based methods}

Following the success of diffusion models in vision \cite{Croitoru-TPAMI-2023}, several generation methods adopted the diffusion modeling framework to generate deepfake audio \cite{Du-AAAI-2024,Huang-IJCAI-2023,Huang-ACMMM-2022,Ju-ArXiv-2024,Shen-ICLR-2024,Tan-TPAMI-2024,Yang-TASLP-2024, Cheng-CVPR-2025b}. Huang \etal~\cite{Huang-IJCAI-2023} present FastDiff-TTS, a conditional diffusion model that follows the architectural design proposed by Ren \etal~\cite{Ren-ICLR-2021}. The authors employ time-aware location variable convolutions for long-term dependency modeling and a noise schedule predictor for sampling acceleration. ProDiff \cite{Huang-ACMMM-2022} is another framework with an architecture inspired by Ren \etal~\cite{Ren-ICLR-2021}, which uses a denoising model with a parametrization that directly predicts the clean data, halving the number of diffusion steps via knowledge distillation. 

The audio encoder/decoder, the phoneme encoder and the duration and pitch predictors proposed by Tan \etal\cite{Tan-TPAMI-2024} are reused in NaturalSpeech 2 \cite{Shen-ICLR-2024}, alongside a diffusion model that learns to predict latent representations conditioned on the input text. 
The encoder/decoder and the duration predictor are further used in NaturalSpeech 3 \cite{Ju-ArXiv-2024}, where, in contrast to previous studies \cite{Tan-TPAMI-2024,Shen-ICLR-2024}, each of the following speech attributes are independently generated by a novel factorized diffusion model: duration, content, prosody and acoustic details. 



\subsection{Multimodal}

\subsubsection{Transformer-based methods}
Recent advancements in talking face generation focus on improving the synchronization of facial movements with speech, while maintaining natural motion and emotional consistency \cite{Jang-CVPR-2024, Cheng-SIGGRAPH-2022, Wang-AAAI-2022, Ling-JSTSP-2023, Zhao-CVPR-2025}. These approaches address challenges such as lip-sync accuracy \cite{Cheng-SIGGRAPH-2022, Wang-AAAI-2022, Ling-JSTSP-2023}, motion stability \cite{Jang-CVPR-2024, Ling-JSTSP-2023} and speaker-specific styles \cite{Jang-CVPR-2024, Cheng-SIGGRAPH-2022}, aiming for realistic human-video synthesis. Jang \etal~\cite{Jang-CVPR-2024} introduce a system that combines talking face generation with TTS, addressing the challenge of generating natural head poses and maintaining consistent speech patterns even with varying facial motions. Their approach leverages a motion sampler and a conditioning method to ensure fluidity in both aspects. Building on the idea of synchronizing audio and visual elements, Cheng \etal~\cite{Cheng-SIGGRAPH-2022} propose VideoReTalking, a method designed to edit real-world talking head videos for perfect lip-sync and emotional consistency. 
In a similar fashion, Wang \etal~\cite{Wang-AAAI-2022} develop a one-shot talking face generation framework. They introduce an audio-visual correlation transformer, which improves lip-sync accuracy by mapping audio to dense motion fields through phoneme and keypoint representations. 
To address emotion-agnostic talking head generation, Gan \etal~\cite{Gan-ICCV-2023} propose emotional adaptation for audio-driven talking-head. The method enhances emotion-agnostic talking-head models by adding three lightweight adaptations: deep emotional prompts, an emotional deformation network, and an emotional adaptation module.

\subsubsection{Diffusion-based methods}

Several frameworks are designed to generate high-quality audio-driven portrait animations, aiming to achieve realism and synchronization \cite{Wei-ArXiv-2024, Wang-ArXiv-2024a, Chen-ArXiv-2024a, Stypulkowski-WACV-2024, Xu-ArXiv-2024a, Xu-ArXiv-2024b, Tian-ArXiv-2024, Ji-CVPR-2025, Li-CVPR-2025a, Wang-CVPR-2025, Guo-CVPR-2025, Liu-CVPR-2025, Xu-CVPR-2025, Hong-ICCV-2025, Siniukov-ICCV-2025}. AniPortrait \cite{Wei-ArXiv-2024} transforms audio into photorealistic animations by extracting 3D facial meshes and head poses, allowing for flexible facial motion editing. Building on this, V-Express \cite{Wang-ArXiv-2024a} focuses on precisely synchronizing lip movements with audio, while maintaining control over facial identity and background through progressive training techniques. EchoMimic \cite{Chen-ArXiv-2024a} offers another solution by integrating audio and facial landmarks using a denoising U-Net architecture, which stabilizes and enhances the natural flow of portrait videos. 
Similarly, Stypu{\l}kowski \etal~\cite{Stypulkowski-WACV-2024} propose an autoregressive diffusion model to achieve realistic talking heads with smooth, expressive movements and accurate lip-sync. 
Xu \etal~\cite{Xu-ArXiv-2024a} also use a diffusion-based framework to improve lip-sync accuracy and motion diversity, employing a hierarchical audio-driven visual synthesis module. 
In a similar direction, VASA \cite{Xu-ArXiv-2024b} produces talking faces, capturing synchronized lip movements and dynamic expressions using a diffusion-based model in a latent facial space, enabling real-time interactions with high realism.
Distinctly, EMO \cite{Tian-ArXiv-2024} generates expressive talking head videos without relying on 3D models, excelling in natural transitions and seamless identity preservation. 

\subsubsection{Other methods}

Recent advancements in talking head generation leverage different models, such as NeRF \cite{Peng-CVPR-2024a, Ye-ArXiv-2023}, Gaussian splatting~\cite{Wang-TVCG-2025, Cho-MM-2024, Li-CVPR-2025}, GANs \cite{Zhang-ICASSP-2022, Doukas-ICCV-2021, Wang-CVPR-2023b}, RNNs \cite{Liu-NeurIPS-2022, Lu-TOG-2021, Gururani-ICCV-2023}, VAEs \cite{Zhang-CVPR-2023b} or CNNs \cite{Hwang-ICASSP-2023}, to address the challenges of synchronization, realism, and efficiency. 
SyncTalk \cite{Peng-CVPR-2024a} is a NeRF-based approach which focuses on speech-driven video generation. SyncTalk enhances synchronization between lip movements, facial expressions and head poses by using a face-sync controller for precise lip-sync, a head-sync stabilizer for natural head movements, and a portrait-sync generator to integrate the generated head with the torso. Similarly, GeneFace++ \cite{Ye-ArXiv-2023} builds on NeRF technology to produce real-time talking face videos with arbitrary speech audio. By improving audio-lip synchronization using pitch contour analysis and incorporating a fast motion-to-video renderer, GeneFace++ offers a robust and efficient solution. To speed up rendering time of the talking videos, several works~\cite{Wang-TVCG-2025, Cho-MM-2024, Li-CVPR-2025} adapt 3D Gaussian Splatting to the task. Notably, GaussianTalker \cite{Cho-MM-2024} integrates the audio and spatial features through cross-attention mechanisms, enabling audio-specific facial changes.

In the context of GANs, Text2Video \cite{Zhang-ICASSP-2022} presents an approach to synthesize videos directly from text, reducing reliance on audio-driven models. Using a phoneme-pose dictionary and a GAN-based architecture, the method achieves high-quality video synthesis with just one minute of training data. 
HeadGAN \cite{Doukas-ICCV-2021} is developed for head reenactment and editing from a single reference image. It integrates 3DMMs for real-time reenactment at approximately 20 FPS. Furthermore, it incorporates audio features for enhanced mouth movement accuracy. Wang \etal~\cite{Wang-CVPR-2023b} introduce TalkLip, a speech-to-lip generation model that enhances lip-speech intelligibility by incorporating a lip-reading expert to penalize incorrect outputs. 

For gesture generation, RNN-based frameworks, such as hierarchical audio-to-gesture (HA2G) \cite{Liu-NeurIPS-2022}, introduce ways to generate co-speech gestures. HA2G extracts multi-level audio features using a hierarchical audio learner. 
Another RNN-based model \cite{Lu-TOG-2021} offers a real-time pipeline to generate personalized photorealistic talking-head animations. This model operates at over 30 FPS and follows a three-stage process: extracting deep audio features, predicting facial dynamics and head motions with an auto-regressive model, and rendering high-fidelity faces through image-to-image translation. 
Some RNN-based methods \cite{Peng-ICCV-2023, Zhong-CVPR-2023} rely on audio-visual cues for realistic face synthesis. Peng \etal~\cite{Peng-ICCV-2023} present a speech-driven 3D face animation model that separates speech content and emotion using an Emotion Disentangling Encoder, while Zhong \etal~\cite{Zhong-CVPR-2023} introduce a two-stage framework for audio-driven person-generic talking face video generation. Both approaches apply RNNs on top of CNN features.

\begin{figure}[!t]
    \centerline{\includegraphics[width=0.85\linewidth]{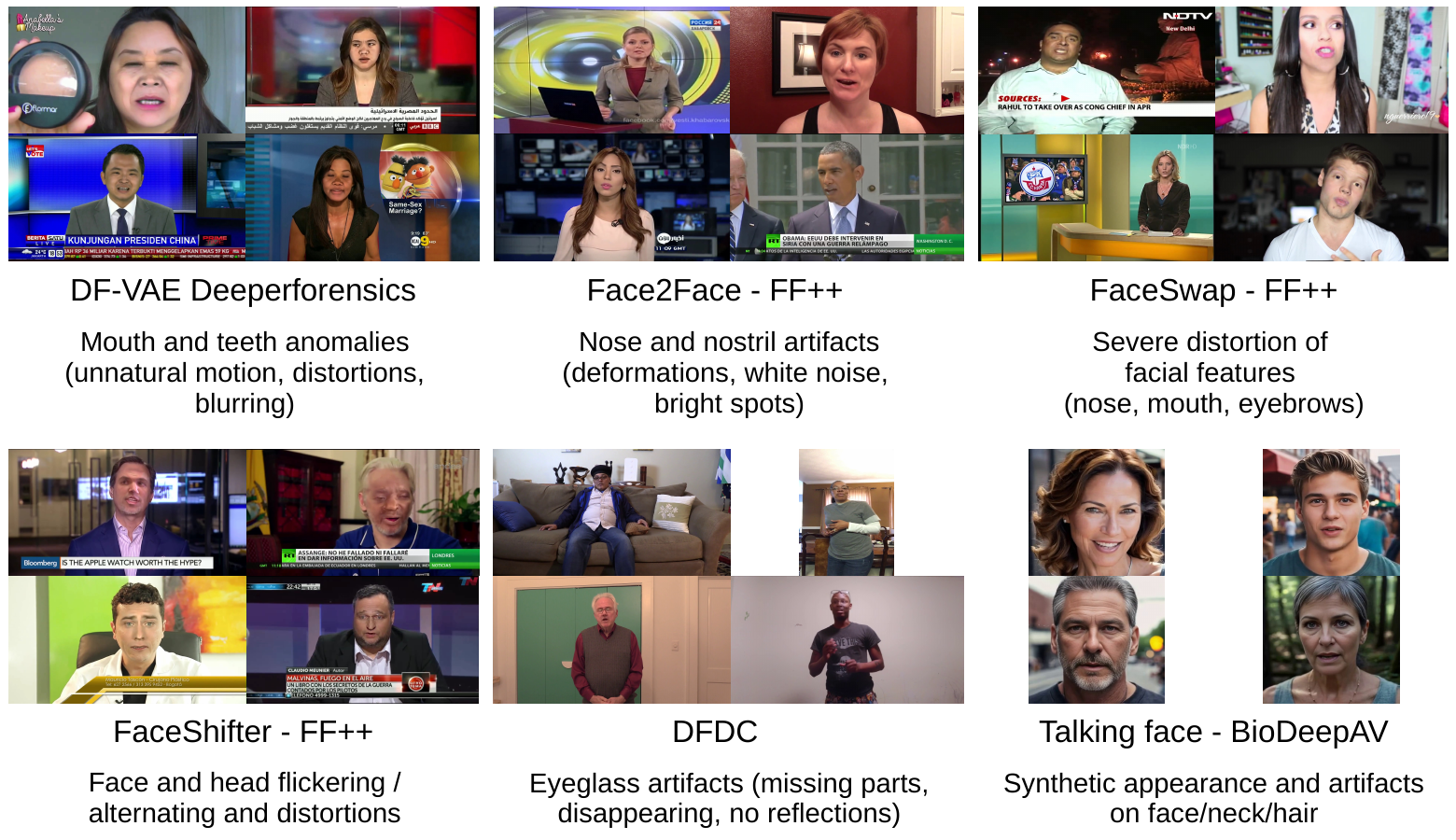}}
    \vspace{-0.3cm}
    \caption{Multiple generative methods from various datasets together with their most common artifacts. Observations are collected from the ExDDV dataset \cite{Hondru-ArXiv-2025}. Best viewed in color.}
    \Description{This figure illustrates failure modes of various deepfake generative methods.}
    \label{fig_deepfake_failures}
    \vspace{-0.3cm}
\end{figure}

\subsection{Comparative analysis of deepfake generation models}

In Figure \ref{fig_deepfake_failures}, we illustrate how different generative methods (collected from multiple datasets) produce various artifacts. DF-VAE~\cite{jiang-CVPR-2020} exhibits motion and blur artifacts around the mouth. Face2face~\cite{thies-CVPR-2016} experiences issues with nose deformations, while FaceSwap~\cite{verdoliva-ICCV-2019} shows severe distortions of various facial features. FaceShifter~\cite{verdoliva-ICCV-2019} is unable to maintain temporal consistency, being affected by flickering. Talking face methods tend to generate faces with unnatural artistic effects. In summary, deepfakes can be discovered via both appearance and motion cues, but specific methods exhibit only a certain subset of the observed artifacts. Next, we analyze the performance and computational speed of GANs versus diffusion models for image generation. For this study, we only look into methods that carried out experiments on datasets with human faces (Celeb-HQ \cite{zhu-ECCV-2022} and FFHQ \cite{Karras-TPAMI-2021}), as well as with higher resolutions (at least $256 \times 256$), as these are more relevant for deepfake media generation. The results are aggregated in Table \ref{tab:gen_method_performance}. According to Haji-Ali \etal~\cite{haji-CVPR-2024}, diffusion models tend to experience difficulties at higher resolution, \eg~creating repetitive elements, stretching objects, etc. Moreover, we observe that the inference time of diffusion models is typically higher. For these reasons, GANs are preferred in high-resolution face image generation. Yet, diffusion models are recognized for their superior conditional generation capabilities \cite{Croitoru-TPAMI-2023}.

\begin{table}[t!]
    \centering
    \caption{Performance vs.~speed comparison of multiple generative methods (both GANs and diffusion models) on Celeb-HQ and FFHQ. The performance is reported in terms of the Fr\'{e}chet Inception Distance (FID). The processing time per image is computed by averaging the duration of the generation process for 50 images on an NVIDIA RTX 4090 GPU with 24GB of VRAM. For diffusion models, we employ a sampler with 50 steps. The asterisk indicates that the respective method focuses on more complicated tasks than just unconditional image generation.}
    \vspace{-0.3cm}
    \setlength\tabcolsep{4.2pt}
    \scriptsize{
    \begin{tabular}{lccccc}
    \toprule
        Class & Model & Resolution & FID on Celeb-HQ & FID on FFHQ & Inference time (s) \\
         \midrule
         \multirow{6}{*}{{GANs}}
         & \cite{Karras-CVPR-2020} & $1024\times 1024$ & - & $2.84$ & $0.3474$ \\
         & \cite{Esser-CVPR-2021} & $256\times 256$ & $10.7$ & $11.4$ & $1.0417$ \\
         & \cite{Karras-NeurIPS-2021} & $1024\times 1024$ & - & $2.79$  & $0.0239$ \\
         & \cite{Karras-TPAMI-2021} & $1024\times 1024$ & $5.06$ & $4.40$  & $0.0888$ \\
         & \cite{Sauer-SIGGRAPH-2022} & $1024\times 1024$ & - & $2.02$ & $0.0871$ \\
         & \cite{Xu-CVPR-2023} & $512\times 512$ & - & $5.7$ & - \\
         \midrule
         \multirow{6}{*}{{Diffusion models}}
         & \cite{Rombach-CVPR-2022} & $256\times 256$ & $5.11$ & $4.98$ & 1.1350 \\
         & \cite{Wang-ArXiv-2024c} & $512\times 512$ & - & $24.92^*$ & $2.7890$ \\
         & \cite{vahdat-NeurIPS-2021} & $256\times 256$ & $7.22$ & - & $2.2865$ \\
         & \cite{kim-ICML-2022} & $256\times 256$ & $7.16$ & - & $3.5458$ \\
         & \cite{liu-AAAI-2024} & $512\times 512$ & $15.51$ & - & $11.3865$ \\
         & \cite{bai-CVPR-2024} & $256\times 256$ & $11.1$ & $18.2$ & $1.3302$ \\
    \bottomrule
    \end{tabular}
    }
    \label{tab:gen_method_performance}
    \vspace{-0.3cm}
\end{table}

\section{Deepfake Detection}

Our second taxonomy, which is illustrated in Figure A2 (in supplementary), comprises deepfake detection methods. The taxonomy clearly indicates that most deepfake detectors are based on CNN architectures. However, with the recent advent of vision and audio transformers, a large body of work on deepfake detection is now based on multi-head attention. To boost detection performance, a considerable number of studies employ hybrid models, combining CNNs with transformers or RNNs, respectively. Less prevalent architectures in deepfake detection are graph and recurrent neural networks. We organize our subsequent presentation of deepfake detection methods according to the input domain. The reviewed articles are further separated according to the employed architectures. We discuss the deepfake localization subtask in Section \ref{sec_supp_B} from the supplementary material.

\subsection{Image}
\subsubsection{CNN-based methods}
Convolutional nets are the most prevalent type of architecture for deepfake detection~\cite{Chen-CVPR-2022, Dong-CVPR-2023, Lin-CVPR-2024b, Yan-CVPR-2024, Tan-CVPR-2024, Nguyen-CVPR-2024, Yao-ICCV-2023, Yan-ICCV-2023, Chen-NeurIPS-2022, Tan-AAAI-2024, Le-AAAI-2024, Kim-CVPRW-2021, Xu-WACVW-2023, Du-CIKM-2019, Huang-CVPR-2023, Shiohara-CVPR-2022, Zhao-CVPR-2021, Dong-ECCV-2022, Le-ICCV-2023,Larue-ICCV-2023, Sun-ICCV-2023, Zhao-ICCV-2021, Hooda-WACV-2024, Ju-WACV-2024, Tantaru-WACV-2024, Trinh-WACV-2021, Ba-AAAI-2024, Yang-AAAI-2022, Nirkin-TPAMI-2022, Lanzino-CVPRW-2024, Ciamarra-WACVW-2024, Jeong-WACV-2022, Cheng-CVPR-2025, Sun-IJCV-2025, Ricker-CVPR-2024, Choi-ArXiv-2024, Li-NeurIPS-2024, Jiaxin-TCSVT-2023, Linbo-TCSVT-2025, Li-TM-2024}. The detection task is commonly formalized as a binary classification, where a CNN backbone is used as feature extractor. The most frequently chosen backbones in this line of work are EfficientNet~\cite{tan-ICML-2019}, XceptionNet~\cite{Chollet-CVPR-2017} and ResNet~\cite{He-CVPR-2015}.

The main direction of research for deepfake detection focuses on developing methods that generalize well across different types of manipulations~\cite{Chen-CVPR-2022, Dong-CVPR-2023, Yan-CVPR-2024, Tan-CVPR-2024, Nguyen-CVPR-2024, Yao-ICCV-2023, Yan-ICCV-2023, Chen-NeurIPS-2022, Tan-AAAI-2024, Le-AAAI-2024, Kim-CVPRW-2021, Xu-WACVW-2023, Du-CIKM-2019, Liu-CVPR-2020, Zhao-CVPR-2021, Jeong-WACV-2022}. To improve generalization, Chen~\etal~\cite{Chen-CVPR-2022} propose an adversarial training pipeline that dynamically identifies which type of forgeries are most challenging for the deepfake detector, and uses them to improve the overall performance. Similarly, Yan~\etal~\cite{Yan-CVPR-2024} increase the diversity and complexity of forgeries, but different from Chen \etal~\cite{Chen-CVPR-2022}, they achieve this through latent space manipulations. Other studies~\cite{Dong-CVPR-2023, Nguyen-CVPR-2024, Yan-ICCV-2023, Tan-AAAI-2024, Le-AAAI-2024, Du-CIKM-2019, Zhao-CVPR-2021} try to identify the common artifacts or features for different types of forgeries. Thus, some methods~\cite{Dong-CVPR-2023, Nguyen-CVPR-2024, Du-CIKM-2019, Zhao-CVPR-2021, Jiaxin-TCSVT-2023} base their solution on local artifact detection and less on the overall identity. Jiaxin~\etal~\cite{Jiaxin-TCSVT-2023} design a two-stream network that fuses multi-scale global features with local inconsistency features obtained by computing the difference between tampered foreground and background images. Yan~\etal~\cite{Yan-ICCV-2023} propose an explicit disentanglement approach using multi-task learning to analyze image information, allowing the detection of features that are shared across various types of forgeries. 

Some studies \cite{Tan-AAAI-2024,Le-AAAI-2024,Luo-CVPR-2021,Qian-ECCV-2020,Jeong-WACV-2022} indicate that the generalization capabilities of deepfake detectors can also be improved by leveraging artifacts in the frequency domain. Moreover, the frequency domain can be leveraged to enhance image watermarking~\cite{Linbo-TCSVT-2025, Li-TM-2024}, which can include adversarial perturbations, serving as defense against forgery models. All these advancements in the generalization of deepfake detectors are validated by Yao~\etal~\cite{Yao-ICCV-2023}, who demonstrate that detectors modeling low-order interactions exhibit superior generalization capabilities.

In contrast to previous works, some deepfake detection methods~\cite{Huang-CVPR-2023, Shiohara-CVPR-2022, Hooda-WACV-2024, Tantaru-WACV-2024} are specialized in identifying specific forgery techniques. Huang~\etal~\cite{Huang-CVPR-2023} and Shiohara~\etal~\cite{Shiohara-CVPR-2022} focus on face swapping. Huang~\etal~\cite{Huang-CVPR-2023} argue that a face-swapped image contains information about the identity present in the target image. Their method builds a face recognition model to detect this identity from a face-swapped image, and leverages the difference between its embedding and the embedding of the source identity to detect deepfakes. Shiohara~\etal~\cite{Shiohara-CVPR-2022} improve face-swapping detection by generating more challenging examples for a face-swapping detector. In this regard, they use face-swapping pipelines where the target and source images are of the same identity and closely resemble each other in terms of face position and other non-identity attributes. In contrast, Hooda~\etal~\cite{Hooda-WACV-2024} and Tantaru~\etal~\cite{Tantaru-WACV-2024} tackle the detection of forgeries in images generated with diffusion models. Hooda~\etal~\cite{Hooda-WACV-2024} detect deepfakes using an ensemble in which the models are using disjoint parts of the input features, aligning with the aforementioned finding of Yao~\etal~\cite{Yao-ICCV-2023}.

Other studies explore deepfake detection methods by examining their fairness~\cite{Ju-WACV-2024, Lin-CVPR-2024b}, level of explainability~\cite{Dong-ECCV-2022, Trinh-WACV-2021}, or level of vulnerability to adversarial attacks~\cite{Hussain-WACV-2021}. Ju~\etal~\cite{Ju-WACV-2024} propose the first approach to tackle fairness in deepfake detectors. Their method uses groups of people specified by the user and ensures that the loss of these groups is similar to each other. 
The method of Ju~\etal~\cite{Ju-WACV-2024} performs well when tested on the same type of forgery as in the training set. Lin~\etal~\cite{Lin-CVPR-2024b} extend the work of Ju~\etal~\cite{Ju-WACV-2024}, aiming to improve generalization across different forgery types. More specifically, their work introduces a disentanglement loss to separate the demographic and forgery specific features. These features are then combined and used in a fairness loss function, which aims to ensure equal importance across different demographic groups. In terms of explainability, Dong~\etal~\cite{Dong-ECCV-2022} try to identify the visual concepts that are relevant for deepfake detectors. Their findings indicate that the features specific to source and target images are, in general, ignored. The focus of the detection models is on visual artifacts. Trinh~\etal~\cite{Trinh-WACV-2021} reinforce this finding by showing that, in addition to visual artifacts, temporal artifacts also serve as evidence for detectors. Hussain~\etal~\cite{Hussain-WACV-2021} show that, despite the progress of deepfake detectors, these models are susceptible to adversarial attacks and future works need to address this drawback.

\subsubsection{GCN-based methods}
As stated before, Yao~\etal~\cite{Yao-ICCV-2023} demonstrate that deepfake detectors with strong generalization capabilities tend to model low-order interactions. Due to their ability to capture such relationships, some studies employ graph convolutional networks (GCNs) for deepfake detection~\cite{Yang-TIFS-2023,Wang-CVPR-2023,Cao-CVPR-2022}. Yang~\etal~\cite{Yang-TIFS-2023} design their graphs with vertices representing features of facial regions and edges capturing the correlations between these regions. They also introduce a masking strategy that removes edges based on their weights. These graphs are then given as input to a GCN, which extracts features for a binary classifier. Wang~\etal~\cite{Wang-CVPR-2023} propose a similar method, but along with the spatial domain features, they also include frequency domain information.

\subsubsection{Transformer-based methods}

State-of-the-art results in a broad range of computer vision tasks are achieved by transformer architectures~\cite{Dosovitskiy-ICLR-2021}. As a result, these architectures are often adopted for deepfake detection~\cite{Dong-CVPR-2022, Wang-ICMR-2022, Aghasanli-ICCVW-2023}. Aghasanli~\etal~\cite{Aghasanli-ICCVW-2023} propose a direct application of transformers for deepfake detection, where the model is used as feature extractor for a binary classifier. The method of Dong~\etal~\cite{Dong-CVPR-2022} is similar to that of Huang~\etal~\cite{Huang-CVPR-2023}, as both approaches harness discrepancies between the explicit identity depicted in the image and the one given by the outer region of the face. However, in contrast to Huang~\etal~\cite{Huang-CVPR-2023}, Dong~\etal~\cite{Dong-CVPR-2022} use the same model for both identities and differentiate between the two regions with two additional tokens. 

\subsubsection{Hybrid architectures}
 
A natural strategy for improving deepfake detection is to combine the aforementioned architectures in a joint pipeline. The primary research focus is on combining transformer and CNN architectures~\cite{Shao-ECCV-2022, Hong-CVPR-2024, Kamat-ICCVW-2023}, though the combination of GANs and CNNs~\cite{Jeong-AAAI-2022} is also explored. For instance, Jeong~\etal~\cite{Jeong-AAAI-2022} train a GAN to generate perturbation maps, which are added to both real and fake images to minimize differences at the frequency level. They argue that using this augmented data to train a CNN-based classifier prevents overfitting to method-specific frequency artifacts, thereby improving the generalization.

Kamat~\etal~\cite{Kamat-ICCVW-2023} focus on exploring different techniques of combining CNN-based and transformer-based feature extractors for deepfake detection. Shao~\etal~\cite{Shao-ECCV-2022} and Hong~\etal~\cite{Hong-CVPR-2024} harness the CNN and transformer combination for a slightly different problem. Their goal is to determine the sequence of facial manipulations used to create a fake image, because deepfakes are frequently created with several manipulation steps. Both methods~\cite{Hong-CVPR-2024, Shao-ECCV-2022} rely on a CNN backbone for feature extraction followed by a transformer that returns the sequence of manipulated regions.

\subsubsection{Traditional machine learning methods}
The earliest approaches to deepfake detection examine the effectiveness of classical machine learning algorithms~\cite{Lugstein-IHMMSec-2021, Guarnera-CVPRW-2020}. Guarnera~\etal~\cite{Guarnera-CVPRW-2020} conjecture that transposed convolutional layers within GANs generate local pixel correlations. Leveraging this, they design an algorithm to extract local features from images, which are then passed to machine learning algorithms such as SVM and k-NN for detection. Lugstein \etal~\cite{Lugstein-IHMMSec-2021} also employ an SVM, but focus on extracting features from the photo response non-uniformity (PRNU) signal. 

\subsection{Video}

\subsubsection{CNN-based methods}

Most deepfake video detection methods are based on plain convolutional models. In the majority of works, 3D convolutions are applied to extract spatio-temporal features from a whole video sequence. Nevertheless, 2D convolutions are also used to extract salient spatial features from individual frames, and then, the results are aggregated to make the final prediction.
Agarwal \etal~\cite{agarwal-WIFS-2020} extract facial features (both static and temporal), and then compare them with a reference set, for each biometric source data, to identify a similar data point, whose label is used for prediction. Similarly, Cozzolino \etal~\cite{cozzolino-ICCV-2021} compare the extracted biometric features from the input video to those of a pristine video.

Some works propose to capture temporal inconsistencies in the video. This is either achieved from successive frames, with some specialized sequential convolutional blocks \cite{gu-AAAI-2022}, or through a hierarchical framework from both local (frame) and global (video snippet) perspectives that can differentiate between real and fake videos \cite{gu-ECCV-2022, zhao-ICICS-2020}.
Based on this objective, some papers focus only on inconsistencies of specific aspects of the face. Haliassos \etal~\cite{haliassos-CVPR-2021, haliassos-CVPR-2022} study mouth movements and propose to learn spatio-temporal representations of mouth motion via two-stage frameworks. Demir \etal~\cite{demir-WACV-2024} magnify the motion of the face and then classify the videos, while also identifying the source generation method.

\subsubsection{Hybrid CNN and RNN architectures}

To obtain a prediction based on the temporal dimension, many detection methods employ a recurrent network to aggregate the latent features extracted by CNNs. Multiple variants of recurrent architectures are used, such as simple RNNs \cite{guera-AVSS-2018, sabir-CVPR-2019}, gated recurrent units (GRUs) \cite{hu-AAAI-2022, Liu-WACV-2023, montserrat-CVPR-2020} and Long Short-Term Memory (LSTM) networks \cite{amerini-ACM-2020, masi-ECCV-2020}. Unlike the rest, Masi \etal~\cite{masi-ECCV-2020} use two branches, each with a different specialization in extracting features, one in the frequency domain and one in the color domain. The resulting representations are aggregated in a bi-directional LSTM, which is optimized via an improved loss function. Montserrat \etal~\cite{montserrat-CVPR-2020} and Liu \etal~\cite{Liu-WACV-2023} extract face crops and employ ArcFace \cite{deng-CVPR-2019} for a better representation of the backbone features. 

\subsubsection{Other hybrid architectures}

Aside from combining CNNs and RNNs, some attempts try to fuse other types of neural networks. An important category is represented by the integration of attention \cite{bonettini-ICPR-2021,wang-AAAI-2023,Yan-ArXiv-2024}. 
Bonettini \etal~\cite{bonettini-ICPR-2021} integrate an attention mechanism into each network in an ensemble of CNNs. Wang \etal~\cite{wang-AAAI-2023} extract noise features from the face crop, as well as a background crop, and feed them into a multi-head attention module. Furthermore, these works \cite{bonettini-ICPR-2021,wang-AAAI-2023} adopt the contrastive learning paradigm in their deepfake video detectors. Different from previous methods, Coccomini \etal~\cite{coccomini-ICIAP-2022} combine various ViTs with an EfficientNet \cite{tan-ICML-2019}, the latter being used for feature extraction.

Due to its demonstrated strength in many tasks, the transformer architecture \cite{vaswani-NeurIPS-2017} is often employed to capture temporal incoherence. For instance, in a number of studies, the transformer is used together with 3D CNNs to extract temporal features \cite{choi-CVPR-2024, Guan-NeurIPS-2022, zheng-ICCV-2021}. Choi \etal~\cite{choi-CVPR-2024} propose a complex framework that utilizes latent features from a pre-trained StyleGAN model \cite{richardson-CVPR-2021}, which are further encoded with GRUs. Xin~\etal~\cite{Xin-TCSVT-2023} employ GRUs in conjuction with SVMs to analyze facial motion features derived from continuous facial landmarks. 
Cai \etal~\cite{cai-CVPR-2023} employ the masked autoencoder pre-training framework to learn facial representations by guiding the masking strategy to focus on hiding face information. The encoder is further fine-tuned on deepfake detection. 

Another interesting direction is to formulate the problem as a graph classification task. Tan \etal~\cite{tan-AAAI-2023} extract embeddings associated with the actions of different facial elements, systematically arrange them in a graph, and then employ a GCN 
to classify the video. Xu \etal~\cite{xu-ICCV-2023} propose a novel strategy: to randomly sample frames from a video and combine them into a single image, called thumbnail. Then, the thumbnail is processed by a Swin Transformer \cite{liu-ICCV-2021} to obtain feature embeddings. Finally, these are fed into a GCN to capture any inconsistency and thus identify fake videos.

\subsection{Audio}

\subsubsection{CNN-based methods}
The ability of CNNs to extract local features allows them to achieve competitive results in spoofed audio detection~\cite{Conti-ICASSP-2022, Zhang-ICLR-2025, Tak-ICASSP-2021, Wang-INTERSPEECH-2023, Hua-SPL-2021}. Tak \etal~\cite{Tak-ICASSP-2021} bring small modifications, such as fixed sinc filters, to the RawNet2 architecture and use it for spoofed speech detection. The same base architecture is further improved by Wang \etal~\cite{Wang-INTERSPEECH-2023} with orthogonal convolutions and temporal convolution networks (TCNs) to enhance the discrimination capability. In contrast, Conti \etal~\cite{Conti-ICASSP-2022} introduce a new pipeline architecture that uses a Speech Emotion Recognition (SER) system as the feature extractor, and a Random Forest as the final classifier. Emotion features are extracted from an intermediate layer of a 3D-Convolutional Recurrent Neural Network.

\subsubsection{GNN-based methods}

Some models use the ability of GNNs to model relationships between entities in order to enhance spoofed speech detection. Graph attention networks (GATs) are used in \cite{Tak-INTERSPEECH-2021,Tak-ASVSPOOF-2021,Jung-ICASSP-2022} to detect artifacts from both temporal and spectral domains. Tak \etal~\cite{Tak-INTERSPEECH-2021} use two separate GATs for relationship modeling between neighboring temporal segments and different sub-bands, respectively, fusing the scores for the final prediction. In a different study \cite{Tak-ASVSPOOF-2021}, the authors use a third GAT to integrate information from temporal and spectral sub-graphs, while Jung \etal~\cite{Jung-ICASSP-2022} propose to combine the two sub-graphs into a single heterogeneous graph via a heterogeneous attention mechanism.
Chen \etal~\cite{Chen-ICASSP-2023} use a GCN to model the relationships from a graph constructed from patches of a spectrogram, outperforming competing models.

\subsubsection{Transformer-based methods}

A growing number of methods use transformers for the synthesized speech detection task \cite{Bartusiak-ACSSC-2021,Martin-ICASSP-2022,Cai-ICASSP-2023,Zhang-IHMMSec-2023,Jung-ICASSP-2022,Liu-ICASSP-2023}. Bartusiak \etal~\cite{Bartusiak-ACSSC-2021} employ a compact convolutional transformer (CCT) to extract feature maps from spectrograms via a convolutional block, concatenate them, and further analyze them via an attention mechanism. 
The CCT is extended in \cite{Bartusiak-ICMLA-2022} to produce a compact attribution transformer (CAT) for the speech synthesizer attribution task, which aims to identify the tool/method that was used to synthesize the speech. 

Mart{\'\i}n-Do{\~n}as \etal~\cite{Martin-ICASSP-2022} present a model trained in a self-supervised manner, employing representations from different transformer layers of a pre-trained wav2vec 2.0 model to detect spoofed speech. 
The wav2vec 2.0 model is also used by Cai \etal~\cite{Cai-ICASSP-2023}, who address partially fake audio detection. They identify fake audio segments by discovering the discontinuity between them. 

Zhang \etal~\cite{Zhang-IHMMSec-2023} aggregate a transformer architecture and a residual network, where the ability of the transformer to model long-term dependencies enables finding correlations between audio frames. 
Rawformer \cite{Liu-ICASSP-2023} aims to improve AASIST \cite{Jung-ICASSP-2022} by replacing the GAT with a transformer encoder. A positional aggregator augments the feature maps obtained by the RawNet2 feature extractor with positional information.

\subsubsection{Other methods}

A method based on monitoring the behavior of neurons from a speaker recognition (SR) model is designed by Wang \etal~\cite{Wang-ACMMM-2020}. The activated neurons from convolutional and fully-connected layers are used as feature vectors in the training process of a shallow network that classifies the input speech as genuine or synthesized.
Zhang \etal~\cite{Zhang-AAAI-2024} introduce Radian Weight Modification (RWM), a continual learning method that adjusts the direction of the gradient based on the means of the intra-class cosine distances of the samples from the current batch.

\subsection{Multimodal}

Multimodal deepfake detection methods typically rely on a fusion strategy to mix information from multiple modalities. We further organize the models according to the underlying architecture. Finally, we provide a comparative analysis of fusion strategies.

\subsubsection{CNN-based methods}

Recent advances in deepfake and multimedia manipulation detection focus on combining audio-visual elements to improve model robustness and accuracy \cite{Raza-CVPR-2023, Cozzolino-CVPR-2023, Kihal-MTA-2023, Mittal-MM-2020, Chugh-MM-2020}. Most of these works~\cite{Chugh-MM-2020, Cozzolino-CVPR-2023, Raza-CVPR-2023} employ ResNet-based frameworks for multimodal feature extraction, detecting deepfakes by analyzing the audio and visual streams either separately~\cite{Cozzolino-CVPR-2023, Chugh-MM-2020} or jointly~\cite{Raza-CVPR-2023}. Mittal \etal~\cite{Mittal-MM-2020} leverage the use of emotions in deepfake detection through audio and video emotion comparison.
Kihal \etal~\cite{Kihal-MTA-2023} introduce VTA-CNN-RF, a deep multimodal spam detection system, achieving over $98\%$ precision in text, image, and audio spam classification using CNNs and Random Forests. 




\subsubsection{Transformer-based methods}

Recent deepfake detection frameworks based on attention mechanisms exploit both audio and visual cues to effectively identify manipulated content \cite{Zhou-ICCV-2021, Oorloff-CVPR-2024, Salvi-JI-2023, Ilyas-ASC-2023, Asha-MS-2024, Liu-SPIC-2023, Feng-CVPR-2023, Zou-ICASSP-2024, Nie-ACMMM-2024, Zhang-ACM-2024, Smeu-CVPR-2025, Kong-CVPR-2025, Yan-CVPR-2025, Han-CVPR-2025, Miao-AAAI-2025, Guo-CVPR-2025b, Wang-CVIU-2024, Klein-MM-2025, Datta-ICCV-2025, Anshul-ICCV-2025, Anshul-WACV-2026}. Many of these works~\cite{Wang-CVIU-2024, Klein-MM-2025, Datta-ICCV-2025, Zhou-ICCV-2021, Oorloff-CVPR-2024, Anshul-ICCV-2025, Anshul-WACV-2026} explore the usefulness of audio-visual synchronization features. Zhou \etal~\cite{Zhou-ICCV-2021} propose a joint audio-visual detection method that leverages the synchronization between modalities, significantly boosting detection accuracy by late-fusing joint predictions with inter-attention mechanisms. Building on this approach, Oorloff \etal~\cite{Oorloff-CVPR-2024} develop a two-stage audio-visual feature fusion method, using contrastive learning and autoencoders in the initial phase to capture audio-visual correspondences, followed by fine-tuning of transformer-based encoders for precise deepfake classification. 
Additional frameworks that employ audio-visual cues have been proposed. For example, Salvi \etal~\cite{Salvi-JI-2023} introduce a framework analyzing audio-visual feature discrepancies over time, uniquely trained on separate monomodal datasets to identify unseen deepfakes. Similarly, AVFakeNet \cite{Ilyas-ASC-2023} is a unified model with dense Swin Transformer modules, which aptly handles variations in facial poses, lighting, and demographic diversity. Asha \etal~\cite{Asha-MS-2024} propose an ensemble-based D-Fence model, utilizing cross-modal attention and self-attenuated neural networks to emphasize correlations between visual and audio elements for improved detection accuracy. Smeu~\etal~\cite{Smeu-CVPR-2025} observe that some deepfake audio-video datasets have spurious features that are exploited by the deepfake detectors, limiting their ability to generalize. To mitigate this bias, they introduce an unsupervised approach that trains a neural network to align the audio and visual features only on real samples.

For both intra and inter modality deepfake detection, Liu \etal~\cite{Liu-SPIC-2023} introduce the Forgery Clues Magnification Transformer (FCMT), which amplifies both intra-modal and cross-modal forgery cues through a distribution difference-based inconsistency computing module. Feng \etal~\cite{Feng-CVPR-2023} tackle audio-visual inconsistencies through an anomaly detection method that trains autoregressive transformers to flag low-probability sequences, using a joint ResNet-18 and VGG-M encoder. Zou \etal~\cite{Zou-ICASSP-2024} advance cross-modality and within-modality regularization by aligning distinct audio and visual signals through multimodal transformers, while Nie \etal~\cite{Nie-ACMMM-2024} introduce FRADE, which relies on adaptive forgery-aware injection and audio-distilled cross-modal interaction to effectively bridge the audio-visual domain gap. Moreover, Yang \etal~\cite{Yang-TIFS-2023b} introduce AVoiD-DF, a model based on a temporal-spatial encoder and a multimodal joint decoder. AVoiD-DF captures inter-modal and intra-modal disharmony, achieving good performance across various forgery techniques.

\subsubsection{Other methods}
Hosler \etal~\cite{Hosler-CVPR-2021} introduce a method for detecting deepfakes by analyzing emotional consistency in human faces and voices using LSTM networks. By predicting emotions from audio and video features, the approach identifies unnatural emotional patterns to flag deepfake media.

\begin{table}[t!]
    \centering
    \caption{Organization of multimodal deepfake detection models according to their fusion strategy.}
    \vspace{-0.3cm}
    \scriptsize{
    \begin{tabular}{lcccc}
    \toprule
         Paper & Early & Middle & Late & Architecture\\
         \midrule
                  Salvi \etal~\cite{Salvi-JI-2023} & \cmark & \cmark & \cmark & Transformer \\
\midrule
         Nie \etal~\cite{Nie-ACMMM-2024} & \cmark & & & Transformer \\
         \midrule
         Raza \etal~\cite{Raza-CVPR-2023} & & \cmark & & CNN \\
         
         Guo \etal~\cite{Guo-CVPR-2025b} & & \cmark & & Transformer \\
         Zhang \etal~\cite{Zhang-MM-2024} & & \cmark & & Transformer \\
         Zhou \etal~\cite{Zhou-ICCV-2021} & & \cmark & & Transformer \\
         Zou \etal~\cite{Zou-ICASSP-2024} & & \cmark & & Transformer \\
         
         Yang \etal~\cite{Yang-TIFS-2023b} & & \cmark & & Hybrid (CNN+Transformer) \\
         \midrule
         Cozzolino \etal~\cite{Cozzolino-CVPR-2023} & & & \cmark & CNN \\
         Kihal \etal~\cite{Kihal-MTA-2023} & & & \cmark & CNN \\
         
         Hosler \etal~\cite{Hosler-CVPR-2021} & & & \cmark & LSTM \\

         Asha \etal~\cite{Asha-MS-2024} & & & \cmark & Transformer \\
         Feng \etal~\cite{Feng-CVPR-2023} & & & \cmark & Transformer \\
         Huang \etal~\cite{Huang-CVPR-2025} & & & \cmark & Transformer \\
         Ilyas \etal~\cite{Ilyas-ASC-2023} & & & \cmark & Transformer \\
         Liu \etal~\cite{Liu-SPIC-2023} & & & \cmark & Transformer \\
         Miao \etal~\cite{Miao-AAAI-2025} & & & \cmark & Transformer \\
         Oorloff \etal~\cite{Oorloff-CVPR-2024} & & & \cmark & Transformer \\
         Smeu \etal~\cite{Smeu-CVPR-2025} & & & \cmark & Transformer \\

                  Goyal \etal~\cite{Goyal-TCSS-2023} & & & \cmark & Hybrid (CNN+LSTM) \\

    \bottomrule
    \end{tabular}
    }
    \label{tab:fusion_taxonomy}
      \vspace{-0.3cm}
\end{table}

\subsubsection{Comparative analysis of fusion strategies for multimodal deepfake detectors}
Multimodal deepfake detection methods typically rely on a fusion strategy to mix information from multiple modalities. We start our analysis by dividing multimodal deepfake detectors into three categories, according to the level at which the fusion is performed: early, middle, late. We present the resulting organization in Table \ref{tab:fusion_taxonomy}. The categorization of fusion strategies reflects a clear preference towards the adoption of late fusion strategies. In late fusion, each modality is processed individually by an independent model, enabling higher interpretability. Hence, an advantage of late fusion strategies is their ability to better accommodate examples with missing modalities, especially with the availability of independent predictions per modality. In contrast, early fusion learns a single joint representation from which the unified prediction is produced. This gives early fusion methods higher capabilities towards integrating multiple modalities and analyzing inter-dependencies across modalities. While early fusion can exploit fine-grained cross-modal correlations, such as subtle audio-visual synchronization artifacts, it does not yield modality-specific predictions, since the modalities are convoluted. Therefore, early fusion strategies are less likely to work when a modality is missing. We further analyze the performance levels reported for various fusion strategies in Section \ref{sec_supp_C} from the supplementary.

\section{Existing Datasets}
\label{sec_experiments}


In Table~\ref{tab:dataset_stats}, we present the most frequently used datasets for deepfake detection, along with the number of real and fake samples, as well as the resolution (for visual datasets) or the bit rate (for audio datasets). We next describe the main steps that are usually employed to build deepfake datasets. The first step of creating a dataset for deepfake detection is collecting the real data. Except for Dolhansky \etal~\cite{dolhansky-arXiv-2020}, who create the original data by recording movies of paid actors, the basic procedure is to scrape the Internet for videos, especially YouTube. Even image datasets use frames extracted from videos.
After acquiring real data, various deepfake methods are applied to generate the fake samples. Due to their excellent trade-off between performance and speed, GANs are adopted for the creation of most datasets, \eg~Celeb-DF~\cite{li-CVPR-2020b}, DeepFake-TIMIT~\cite{korshunov-arXiv-2018}, DFFD~\cite{dang-CVPR-2020}, FakeSpotter~\cite{wang-IJCAI-2019} and FakeAVCeleb~\cite{khalid-NeurIPS-2021}. VAEs represent the method of choice only for a few datasets, \eg~DeeperForensics~\cite{jiang-CVPR-2020} and ASVspoof 2019-LA~\cite{wang-CSL-2020}. Diffusion models are recent and powerful generative methods, yet they require more computation. Hence, only a couple of recent datasets employ them to create the fake samples, \eg~GenVideo~\cite{chen-arXiv-2024c} and DiffusionFace~\cite{Chen-ArXiv-2024d}. Several datasets, especially the audio ones, are generated by using more than one method, \eg~GenVideo~\cite{chen-arXiv-2024c}, ForgeryNet~\cite{He-CVPR-2021}, FaceForensics++~\cite{verdoliva-ICCV-2019}, DFDC~\cite{dolhansky-arXiv-2020}, LAV-DF~\cite{cai-DICTA-2022}, WaveFake~\cite{frank-ArXiv-2021}, ASVspoof 2019-LA~\cite{wang-CSL-2020}, ASVspoof 2021-LA/DF~\cite{yamagishi-ASVspoof-2021}, FoR~\cite{reimao-SpeD-2019} and MLAAD~\cite{muller-IJCNN-2024}. 
A few visual datasets \cite{verdoliva-ICCV-2019, chen-arXiv-2024c} rely on online tools to create deepfakes, the most popular tool being FaceSwap\footnote{\href{https://github.com/deepfakes/faceswap}{https://github.com/deepfakes/faceswap}}. 

Depending on the input modality, different metrics are commonly reported. For the visual modalities, the most frequent metric is the area under the curve (AUC). Given that deepfake detection is a binary classification task, the AUC score can illustrate the ability of the model to differentiate between real and fake samples. The AUC is obtained by plotting the True Positive Rate against the False Positive Rate for multiple thresholds, and then computing the area under the resulting curve. Accuracy is an alternative metric that can be used to assess the overall performance of a deepfake detection model. Nevertheless, deepfake detection datasets are usually imbalanced, making accuracy a less preferred option. For the audio modality, models are regularly evaluated via the equal error rate (EER). Its popularity is given by the robustness to class imbalance, while equally assessing false positives and false negatives. EER is computed by finding the intersection of the False Acceptance Rate and the False Rejection Rate. 

We report detailed quantitative results on the most popular benchmarks in Section \ref{sec_supp_D} from the supplementary material. With few exceptions, the state-of-the-art results indicate that current datasets are saturated. Essentially, this happens because detectors are typically fed with training samples from the same set of deepfake generation methods that are used during testing. To alleviate this issue, we further propose a new evaluation benchmark.

\begin{table}[t!]
    \centering
    \caption{Datasets that are commonly used in deepfake detection literature, separated by domain. AV stands for audio-video (multimodal).}
     \vspace{-0.3cm}
\scriptsize{
    \begin{tabular}{clrrc}
    \toprule
         Modality & \multirow{1}{*}{Dataset} & \multirow{1}{*}{\#Real} & \multirow{1}{*}{\#Fake} & \multirow{1}{*}{Resolution/frequency} \\
         \midrule
         \multirow{4}{*}{{Image}} 
         & DFFD~\cite{dang-CVPR-2020} & 58,703 & 240,336 & 250$\times$250 - 1024$\times$1024 \\
         & FakeSpotter~\cite{wang-IJCAI-2019} & 6,000 & 6,000 &  224$\times$224 \\
         & ForgeryNet~\cite{He-CVPR-2021} &  1,438,201 & 1,457,861 & 240$\times$240 - 1080$\times$1080 \\
         & DiffusionFace~\cite{Chen-ArXiv-2024d} & 30,000 & 600,000 & 256$\times$256 \\

         \hline 

         \multirow{7}{*}{{Video}}
         & FaceForensics++~\cite{verdoliva-ICCV-2019} & 1,000  & 4,000 & 512$\times$512 \\
         & DeeperForensics~\cite{jiang-CVPR-2020} & 48,475  & 11,000 & 1920$\times$1080 \\
         & Celeb-DF~\cite{li-CVPR-2020b} & 590 & 5,639 & 256$\times$256\\
         & WildDeepfake~\cite{zi-ACM-2020} & 3,805 & 3,509 & varying\\
         & DeepFake-TIMIT~\cite{korshunov-arXiv-2018} & 0 & 620 & 64$\times$64/128$\times$128 \\
         & UADFV~\cite{yang-ICASSP-2019} & 98 & 98 & 294$\times$500 \\
         & GenVideo~\cite{chen-arXiv-2024c} & 1,224,511 & 1,089,671 & 224$\times$224 - 1280$\times$2048\\
        \hline 
         
        \multirow{9}{*}{{Audio}} 
        & WaveFake~\cite{frank-ArXiv-2021} & 0 & 117,985 & 16 kHz\\
        & ASVspoof 2019-LA~\cite{wang-CSL-2020} & 12,483 & 108,978 & 16 kHz\\
        & ASVspoof 2021-LA~\cite{yamagishi-ASVspoof-2021} & 16492 & 148148 & 16 kHz \\
        & ASVspoof 2021-DF~\cite{yamagishi-ASVspoof-2021} & 20,637 & 572,616 & 16 kHz \\
        & In-the-Wild~\cite{muller-INTERSPEECH-2022} & 19,963 & 11,816 & 16 kHz \\
        & ADD 2022~\cite{yi-ICASSP-2022} & 91,464 & 358,082 & 16 kHz \\
        & ADD 2023~\cite{yi-add-ArXiv-2023} & 243,194 & 273,874 & 16 kHz \\
        & FoR~\cite{reimao-SpeD-2019} & 111,000 & 87,285 & 16 kHz \\
        & MLAAD~\cite{muller-IJCNN-2024} & 0 & 154,000 & 22 kHz \\

        \hline

        \multirow{4}{*}{{AV}}
         & FakeAVCeleb~\cite{khalid-NeurIPS-2021} & 500 & 19,500 & 224$\times$224 \\
         & LAV-DF~\cite{cai-DICTA-2022} & 36,431 & 99,873 & 224$\times$224 \\
         & DFDC~\cite{dolhansky-arXiv-2020} & 23,654 & 104,500 & 1920$\times$1080/1080$\times$1920 \\ 
         & KoDF~\cite{Kwon-ICCV-2021} & 62,166 & 175,776 & 512$\times$512\\

        \bottomrule
    \end{tabular}
    }
    \vspace{-0.3cm}
    \label{tab:dataset_stats}
\end{table}



\section{Proposed Benchmark}
We create a new dataset, called BioDeepAV\footnote{Available at: \href{https://github.com/CroitoruAlin/biodeep}{https://github.com/CroitoruAlin/biodeep}}, to assess the out-of-domain generalization capabilities of deepfake detection models. Our primary focus is on generating videos featuring talking faces, but we also include audio-video examples with audio-only manipulations. Figure A8 (in supplementary) depicts a few frames from various deepfake videos in BioDeepAV.

\paragraph{Generated Data.} We generate over 1,600 deepfake videos using four recent methods specialized in talking-face synthesis~\cite{Liu-ArXiv-2024, Chen-ArXiv-2024a, Peng-CVPR-2024a, Cho-MM-2024}. These approaches base their solutions on the recent development of diffusion models~\cite{Liu-ArXiv-2024, Chen-ArXiv-2024a}, NeRFs~\cite{Peng-CVPR-2024a} and Gaussian Splatting~\cite{Cho-MM-2024}. We use three face image sources to sample target identities. First, we create 300 synthetic faces using RealVisXLv5\footnote{\href{https://civitai.com/models/139562/realvisxl-v50}{https://civitai.com/models/139562/realvisxl-v50}} and supplement these with faces from the LAION-Face~\cite{Zheng-CVPR-2022} and HDTF~\cite{Zhang-CVPR-2021} datasets. Three of the methods~\cite{Chen-ArXiv-2024a, Peng-CVPR-2024a, Cho-MM-2024} also require head motion information as a conditioning signal, which we obtain from the HDTF~\cite{Zhang-CVPR-2021} dataset. In addition to face images and motion cues, all methods also require an audio file to condition their talking-face generation. For this, we use the samples from a dataset of English dialects~\cite{Demirsahin-LREC-2020}, the audio from the HDTF~\cite{Zhang-CVPR-2021} dataset, and over 700 deepfake audio samples created by us. To generate these synthetic audio samples, we employ StyleTTS~\cite{Li-NeurIPS-2023}, SSR-Speech~\cite{Wang-arXiv-2024f} and YourTTS~\cite{Casanova-ICML-2022}, which support both text-to-speech synthesis~\cite{Li-NeurIPS-2023, Wang-arXiv-2024f, Casanova-ICML-2022} and voice conversion~\cite{Casanova-ICML-2022}. We source text prompts for text-to-speech synthesis from the LibriTTS dataset~\cite{Zen-Interspeech-2019}, and use the speakers from this dataset for voice conversion, with target audio sourced from the dataset of English dialects~\cite{Demirsahin-LREC-2020}.

\begin{table}[t!]
    \centering
    \caption{Results (in terms of AUC) of four state-of-the-art deepfake detectors on the original test sets versus BioDeepAV. UCF~\cite{Yan-ICCV-2023}, RECCE~\cite{Cao-CVPR-2022}, TALL~\cite{xu-ICCV-2023}, F3Net~\cite{Qian-ECCV-2020}, StA~\cite{Yan-ArXiv-2024} and XceptionNet~\cite{verdoliva-ICCV-2019} are originally tested on FaceForensics++~\cite{verdoliva-ICCV-2019}, while MRDF is originally tested on FakeAVCeleb~\cite{khalid-NeurIPS-2021}.}
    \vspace{-0.3cm}
    \scriptsize{
    \begin{tabular}{lccc}
    \toprule
        Method & Venue & Original Test & BioDeepAV \\
        \midrule
        StA~\cite{Yan-ArXiv-2024} & ArXiv 2024 & 0.9420 & 0.6195\\
        XceptionNet~\cite{verdoliva-ICCV-2019} & ICCV 2019 & 0.9637 & 0.5677 \\
        F3Net~\cite{Qian-ECCV-2020} & ECCV 2020 & 0.9449 & 0.5010 \\
        RECCE~\cite{Cao-CVPR-2022} & CVPR 2022 & 0.9422 & 0.5001 \\
        TALL~\cite{xu-ICCV-2023} & ICCV 2023 & 0.9987 &  0.4935\\
        UCF~\cite{Yan-ICCV-2023} & ICCV 2023 & 0.9527 &  0.4882\\
        \hline
        MRDF~\cite{Zou-ICASSP-2024} & ICASSP 2024 &  0.9243 &  0.5852\\
        \bottomrule
    \end{tabular}
    }
    \label{tab:benchmark results}
        \vspace{-0.3cm}
\end{table}

\paragraph{Real Data.} We sample real videos for our experiments from two datasets, HDTF~\cite{Zhang-CVPR-2021} and TalkingHead-1KH~\cite{Wang-CVPR-2021}. We include all available videos from HDTF, while sampling an additional 2,000 videos from TalkingHead-1KH.

\paragraph{Experiments.} We run the experiments using the DeepfakeBench benchmark~\cite{Yan-NeurIPS-2023}, which implements state-of-the-art deepfake detectors. For our analysis, we choose three image-based detectors, namely UCF~\cite{Yan-ICCV-2023}, RECCE~\cite{Cao-CVPR-2022} and a model based on XceptionNet~\cite{verdoliva-ICCV-2019}, one detector applied on the frequency domain, namely F3Net~\cite{Qian-ECCV-2020}, two video-based detectors, namely TALL~\cite{xu-ICCV-2023} and StA~\cite{Yan-ArXiv-2024}, and one audio-visual detector, namely MRDF~\cite{Zou-ICASSP-2024}. MRDF is not implemented in the DeepFakeBench benchmark, so we employ the official implementation in our experiments. UCF~\cite{Yan-ICCV-2023}, RECCE~\cite{Cao-CVPR-2022}, TALL~\cite{xu-ICCV-2023}, F3Net~\cite{Qian-ECCV-2020}, StA~\cite{Yan-ArXiv-2024} and XceptionNet~\cite{verdoliva-ICCV-2019} are trained on FaceForensics++~\cite{verdoliva-ICCV-2019}, while MRDF~\cite{Zou-ICASSP-2024} is trained on FakeAVCeleb~\cite{khalid-NeurIPS-2021}. In Table~\ref{tab:benchmark results}, we report the video AUC of these detectors on BioDeepAV and their original test sets, respectively. The considered methods always surpass the 90\% threshold when tested in-domain, yet all methods register drastic performance drops (higher than 30\%) on BioDeepAV.
The findings clearly demonstrate that current detectors struggle to identify the authenticity of talking faces generated by the novel (unseen) generative models included in BioDeepAV, highlighting the need for further research in this area. To strengthen this observation, we analyze the performance of deepfake detectors per generative method, in Section \ref{sec_supp_E} from the supplementary. 

\section{Conclusions and Future Perspectives}
\label{sec_conclusion}

In this paper, we reviewed deepfake generation and detection methods, constructing a comprehensive taxonomy of methods across image, video, audio and multimodal domains. After discussing the methods included in our taxonomy, we turned our attention to datasets used for deepfake detection, with a particular focus on the results reported by top performing models. Moreover, we evaluated some of the best methods on our novel benchmark, BioDeepAV, aiming to assess the generalization capacity of current deepfake detectors to out-of-distribution data. The results show that the distribution gap can greatly affect state-of-the-art deepfake detectors, pinpointing the need for more robust models.

\subsection{Future Perspectives} As the deepfake generation technology continues to evolve, a number of complex challenges are raised, spanning from a technical nature to ethical and societal concerns. Furthermore, based on the observed gaps in deepfake literature, there are several directions which we recommend exploring in future work. We next split the discussion between deepfake generation and detection.

\paragraph{Generation safeguarding.}
The widespread of publicly available generation methods, together with their increased capability, has accelerated the accessibility and realism of synthetic media, making it critical to reflect on the potential risks. First of all, researchers should develop safeguards (\eg~invisible watermarks), while the developers of open-source models should integrate such mechanisms or implement usage restrictions to mitigate misuse. In parallel with the development of generative methods, regulatory and legal frameworks need to be implemented to define and enforce boundaries of fair and safe use. Finally, but not least important, the resilience of the society against manipulated media can be boosted through public awareness campaigns and digital literacy initiatives.

\paragraph{Cross-domain and open-set benchmarks.}
We consider that the most important future direction is the development of deepfake detectors that can generalize across multiple generative tools. Our new benchmark, BioDeepAV, will come in handy to test the generalization capacity of deepfake detection models in the future. Cross-domain and open-set benchmark need to be continuously developed, as generative methods get better over time.

\paragraph{One-class learning.}
We foresee that the principal avenues for improving generalization in deepfake detection lie in the exploration of multimodal architectures, while shifting away from traditional supervised learning. Although, unsupervised methods have been tried in current literature, they still lag behind supervised methods. One field from which future research can draw inspiration for improvement is video anomaly detection, where models are trained exclusively on normal data on various pretext tasks \cite{Barbalau-CVIU-2023}. Future deepfake detection frameworks could adopt a one-class classification paradigm, utilizing self-supervised pretext tasks specifically designed to capture the  spatio-temporal dynamics of real talking faces. In this setting, models are trained solely on real samples, and, at inference time, the degree of discrepancy between the input sample and the learned distribution of real data serves as a metric for identifying forgeries. 

\paragraph{Explainable detection.}
Another area that is not sufficiently explored is the development of explainable deepfake detectors \cite{Hondru-ArXiv-2025}. Knowing when and why deepfake detectors fail is an important aspect for making deepfake detectors more user friendly, but this has often been disregarded. While current detectors mostly rely on deep neural networks, an important downside of such models is that they are unable to quantify their uncertainty. To this end, studying approaches to calibrate deepfake detectors could lead to the development of enhanced models able to indicate when their prediction is uncertain.

\paragraph{Person-of-interest deepfake detection.}
An understudied line of work in deepfake detection is the relatively recent person-of-interest (POI) deepfake detection \cite{Cozzolino-CVPR-2023,Tian-TCVST-2023}. POI deepfake detection focuses on verifying whether a known individual, \eg~a public figure, is authentically represented in a media file, rather than trying to detect any possible deepfake \cite{Cozzolino-CVPR-2023,Salvi-ICCV-2025}. This shift brings several practical advantages. First, it allows the use of identity-specific signals. Because the system is tested on a particular person, it can learn fine-grained biometric and behavior patterns, \eg facial micro-expressions, voice idiosyncrasies, blinking rhythms, or speech cadence. This typically results in higher accuracy and robustness when analyzing content involving that individual. Second, POI detection can be less vulnerable to generalization failures. Standard deepfake detectors often struggle with new synthesis methods or unseen datasets, because they attempt to model fake data in a broad sense. In contrast, POI detection systems operate in a matching problem setup, which is a more stable and constrained problem. Despite these advantages, there is only one available dataset for POI deepfake detection~\cite{Tian-TCVST-2023}. In future work, we advocate towards the development of both benchmarks and methods for POI deepfake detection. Developments should consider updating POI detection systems via continual learning, as new real footage becomes available, adapting to aging, hairstyle changes, etc. In this context, robustness to poisoning attacks is of utmost importance.

\paragraph{Multimedia blockchain.} Propose a blockchain solution for multimedia sharing, which aims to provide a mechanism to only share verified content and to precisely determine the source of shared multimedia content, thereby circumventing deepfake detection. We imagine a media ecosystem where every photo, video, or audio clip is ``born verified''. When a journalist or a person captures content, their device cryptographically signs the file at the moment of creation using a secure hardware key. A non-fungible token (NFT) is immediately recorded on a blockchain. The NFT is a unique digital identifier that exactly represents the created piece of media, embedding metadata such as creator identity, time, location, and device signature. When a media outlet publishes or shares the content, it does so by transferring or referencing the NFT. Any recipient, another newsroom or end user, can trace the NFT on-chain to confirm provenance and integrity. If even a single pixel or frame is altered, the hash no longer matches the original record, signaling tampering. Rather than trying to detect whether content is fake, the proposed system flips the paradigm: only content with a verifiable origin and unbroken chain of custody is trusted. Over time, platforms and audiences should rely on this new proof-of-authenticity standard. Thus, trust becomes a property of the infrastructure rather than a probabilistic judgment. Content without a valid NFT or with broken provenance is treated as unverified by default. While the proposed multimedia blockchain reduces the need for deepfake detection systems, there several practical challenges that have to be addressed. First, it requires widespread adoption across device manufacturers, journalists, and platforms, as well as secure hardware to prevent key compromise at the source. Privacy concerns may arise when embedding metadata like location. Scalability and storage constraints must be addressed, since blockchains cannot hold large media files directly. Finally, governance and standardization are critical. Without global agreement on protocols and trust anchors, fragmentation could undermine the very authenticity the system aims to guarantee.

\begin{acks}
This work was supported by a grant of the Ministry of Research, Innovation and Digitization, CCCDI - UEFISCDI, project number PN-IV-P6-6.3-SOL-2024-2-0227, within PNCDI IV.

\end{acks}

\bibliographystyle{ACM-Reference-Format}
\bibliography{veryshortref-nopages}

\section{Brief Supplementary Overview}

In the supplementary, we explain in detail how the most popular and interesting generative developments work, we present results on the most popular deepfake detection benchmarks, and we discuss the advancements made in facial forgery localization.

\begin{figure}[!t]
\begin{center}
\centerline{\includegraphics[width=1.0\linewidth]{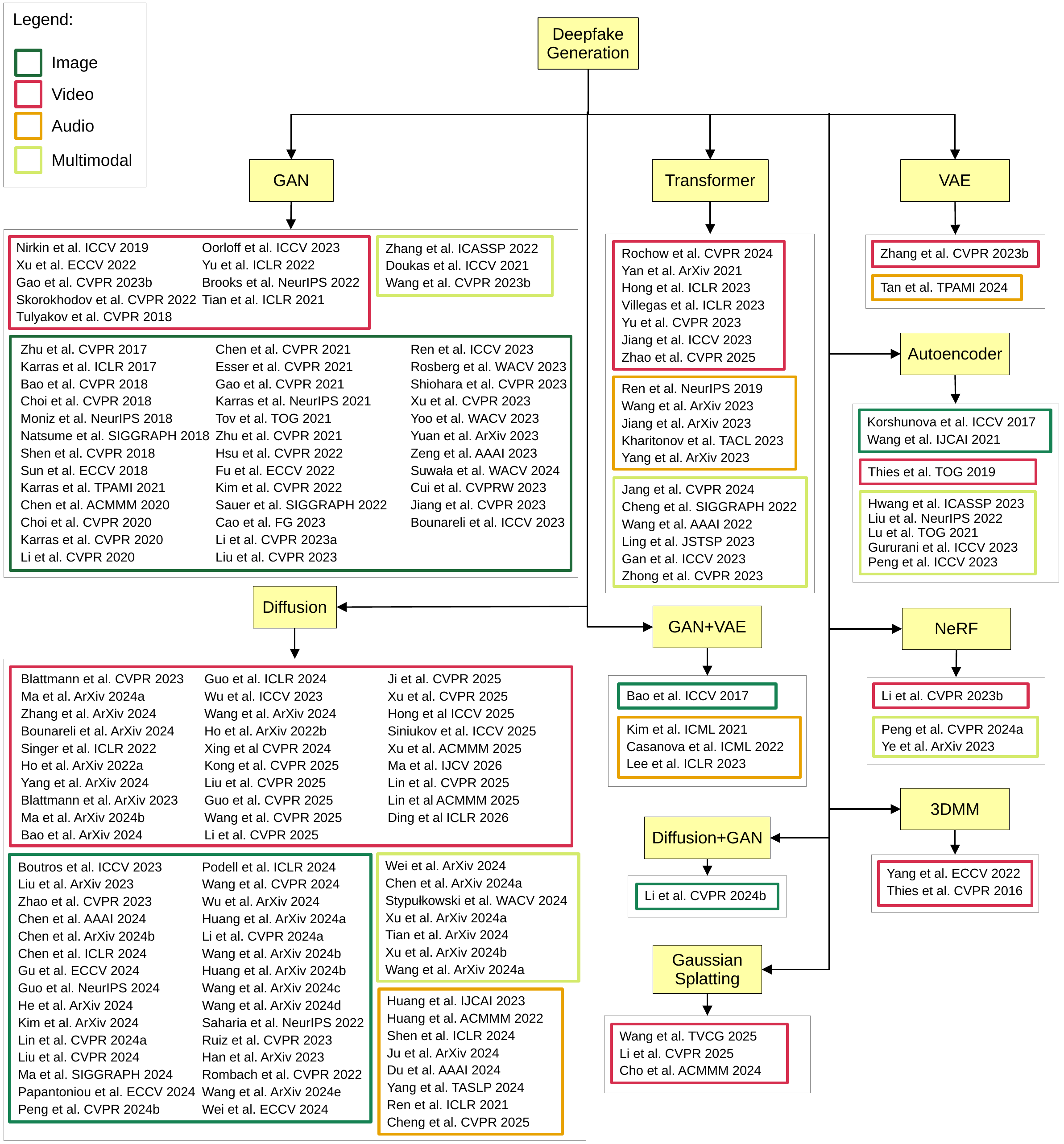}}
\vspace{-0.3cm}
\caption{A taxonomy of the state-of-the-art deepfake generation methods. The methods are divided into different kinds of generative architectures. In the ``transformer'' category, we included the methods that use the transformer architecture for autoregressive or masked modeling-based generation. For each architecture, we separate the methods based on the media types. Large groups are further divided according to the deepfake types presented in Section \ref{sec_deepfake_types}. References are clickable links to papers. Best viewed in color.}
\Description{A taxonomy of the state-of-the-art deepfake generation methods. The methods are divided into different kinds of architectures. For each architecture, we separate the methods based on the media types. Large groups are further divided according to the deepfake types presented in Section \ref{sec_deepfake_types}.}
\label{fig_taxonomy_gen}
\vspace{-0.65cm}
\end{center}
\end{figure}

\begin{figure}[!t]
\begin{center}
\centerline{\includegraphics[width=0.88\linewidth]{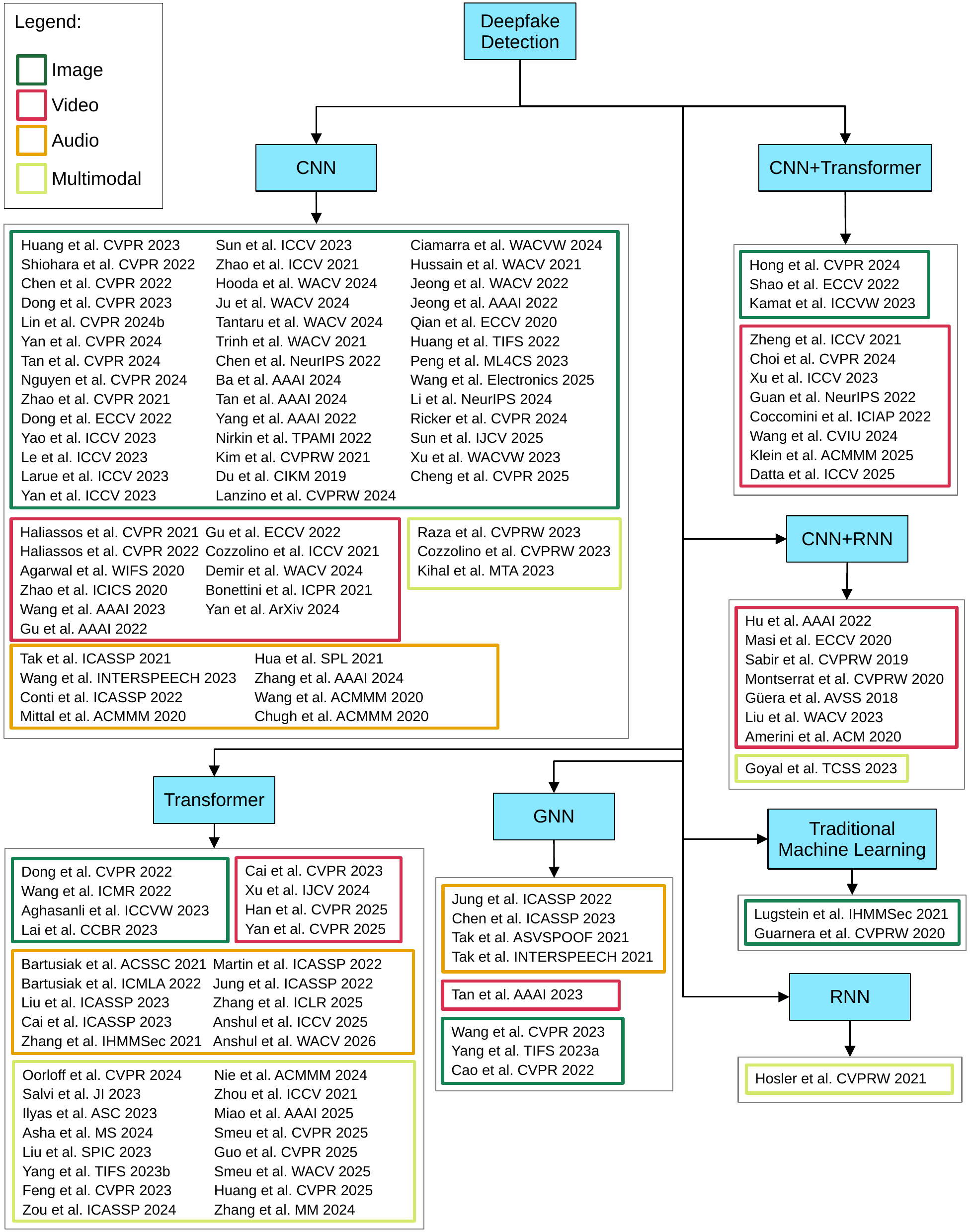}}
\vspace{-0.3cm}
\caption{A taxonomy of the state-of-the-art deepfake detection methods. The methods are divided into different kinds of architectures. For each architecture, we separate the methods based on the media types. Large groups are further divided according to the deepfake types presented in Section \ref{sec_deepfake_types}. References are clickable links to papers. Best viewed in color.}
\Description{A taxonomy of the state-of-the-art deepfake detection methods. The methods are divided into different kinds of architectures. For each architecture, we separate the methods based on the media types. Large groups are further divided according to the deepfake types presented in Section \ref{sec_deepfake_types}.}
\label{fig_taxonomy_det}
\vspace{-0.8cm}
\end{center}
\end{figure}

\section{Usage Examples for Mainstream Generative Frameworks}
\label{sec_supp_A}

\begin{figure}[t]
\begin{center}
\centerline{\includegraphics[width=0.7\linewidth]{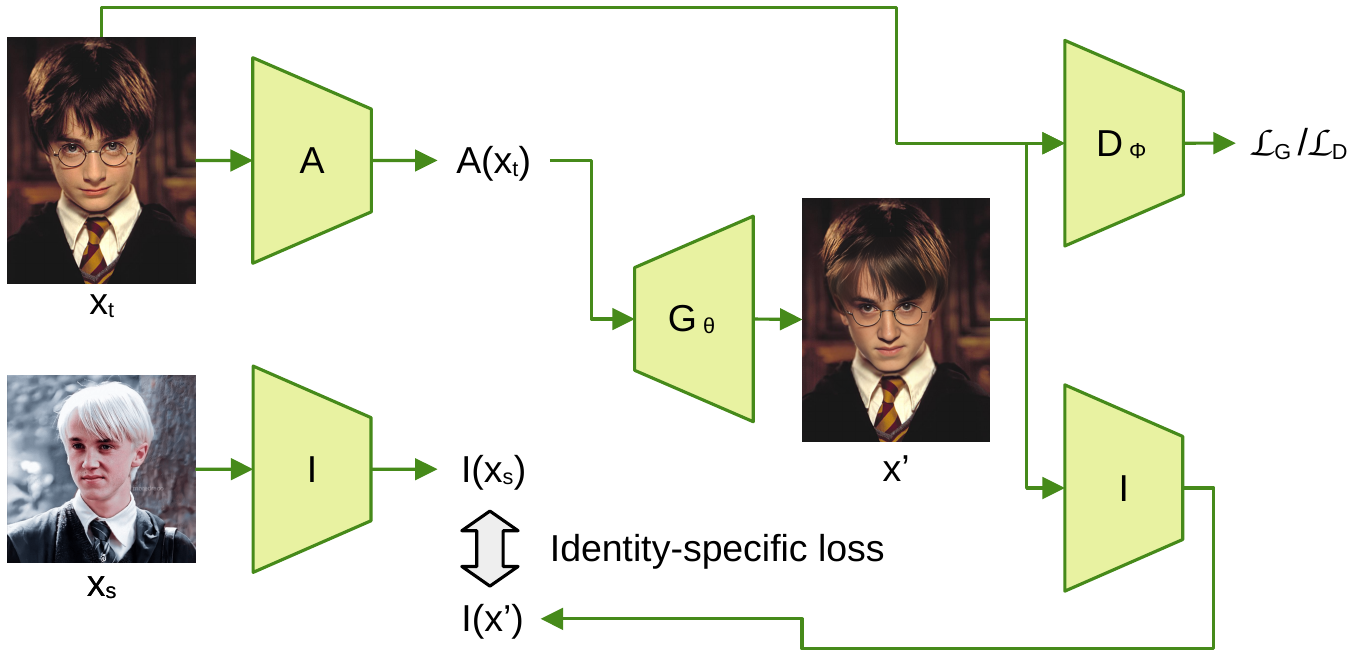}}
\vspace{-0.4cm}
\caption{An overview of face swapping based on GANs. The generative process is conditioned on an identity encoder $I$ and an attribute encoder $A$, aiming to preserve the target attributes while replacing the target identity with the source identity.}
\Description{An overview of face swapping based on GANs. The generative process is conditioned on an identity encoder $I$ and an attribute encoder $A$, aiming to preserve the target attributes while replacing the target identity with the source identity.}
\label{fig_gans_deepfake}
\vspace{-0.5cm}
\end{center}
\end{figure}

\subsection{GAN Usage Example}
In Figure \ref{fig_gans_deepfake}, we illustrate a typical face swapping pipeline powered by GANs. The generator $ G_\theta $ is conditioned on features derived from two sources. The identity encoder $I$ extracts features from the identity source image $x_s$, while the attribute encoder $A$ extracts features from the target image $x_t$. Commonly, the encoder $I$ is a (pre-trained) face recognition model, while $A$ is a randomly initialized encoder trained along with the rest of the pipeline. The generator harnesses the input features to produce an image $x'$ that retains the attributes of $x_t$, while swapping the target identity with the source identity from $x_s$. Aside from the previously discussed adversarial losses, the pipeline incorporates a reconstruction loss between $x_t$ and $x'$ to ensure attribute preservation. Additionally, an identity-specific loss is employed, typically calculated as the cosine similarity between the feature vectors extracted by the identity encoder $I$ from $x'$ and $x_s$.

\begin{figure}[t]
\begin{center}
\centerline{\includegraphics[width=0.66\linewidth]{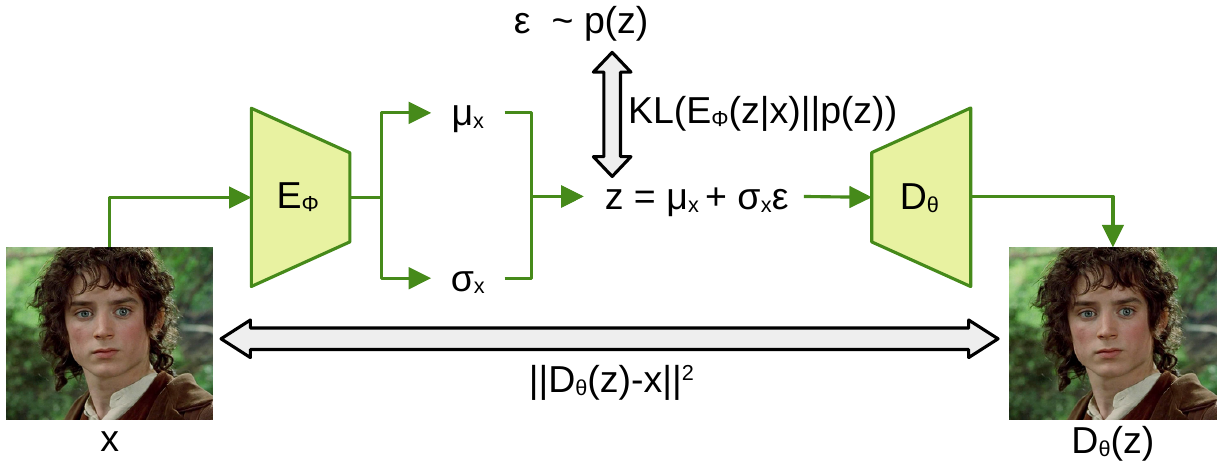}}
\vspace{-0.4cm}
\caption{An overview of face synthesis based on VAEs. The KL divergence is used to minimize the distribution gap between the distribution of $z$ and the standard Gaussian distribution $p(z)$.}
\Description{An overview of face synthesis based on VAEs. The KL divergence is used to minimize the distribution gap between the distribution of $z$ and the standard Gaussian distribution $p(z)$.}
\label{fig_vaes_deepfake}
\vspace{-0.5cm}
\end{center}
\end{figure}

\subsection{VAE Usage Example}

In Figure \ref{fig_vaes_deepfake}, we showcase the training process of a VAE for face synthesis. To enable gradient back-propagation through the encoder, the reparameterization trick is applied to sample $z\sim\mathcal{N}(\mu_x, \sigma_x\mathbf{I})$. During inference, $z$ is directly sampled from $p(z)$ and fed into the decoder.

\begin{figure}[t]
\begin{center}
\centerline{\includegraphics[width=0.82\linewidth]{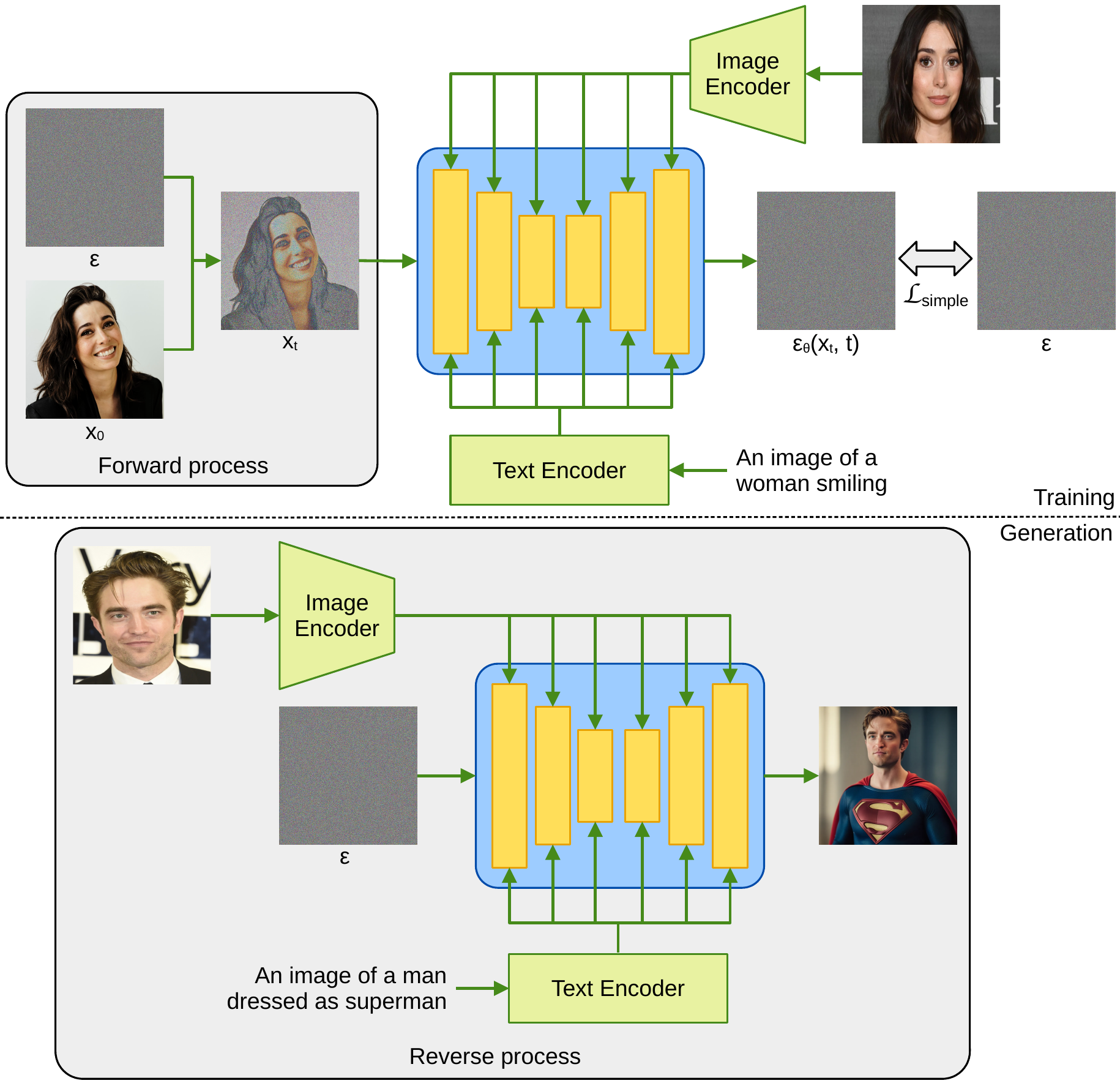}}
\vspace{-0.4cm}
\caption{An overview of text-conditional personalized generation based on diffusion models. The model aims to generate images of a source identity conditioned on a text prompt.}
\Description{An overview of text-conditional personalized generation based on diffusion models. The model aims to generate images of a source identity conditioned on a text prompt.}
\label{fig_diffusion_deepfake}
\vspace{-0.5cm}
\end{center}
\end{figure}

\subsection{Diffusion Usage Example}

In Figure \ref{fig_diffusion_deepfake}, we illustrate the training and inference processes of a text-conditional personalized generative pipeline for a given identity. During training, pairs of images representing the same identity are used. One image undergoes the forward process, and the model is tasked with predicting the added noise. The second image in the pair, along with a text description of the first image, serve as conditional inputs to guide the model. The generative (reverse) process begins with standard Gaussian noise and progressively generates an image that aligns with the provided text description, while preserving the identity from the given source image.

\begin{figure}[t]
\begin{center}
\centerline{\includegraphics[width=0.75\linewidth]{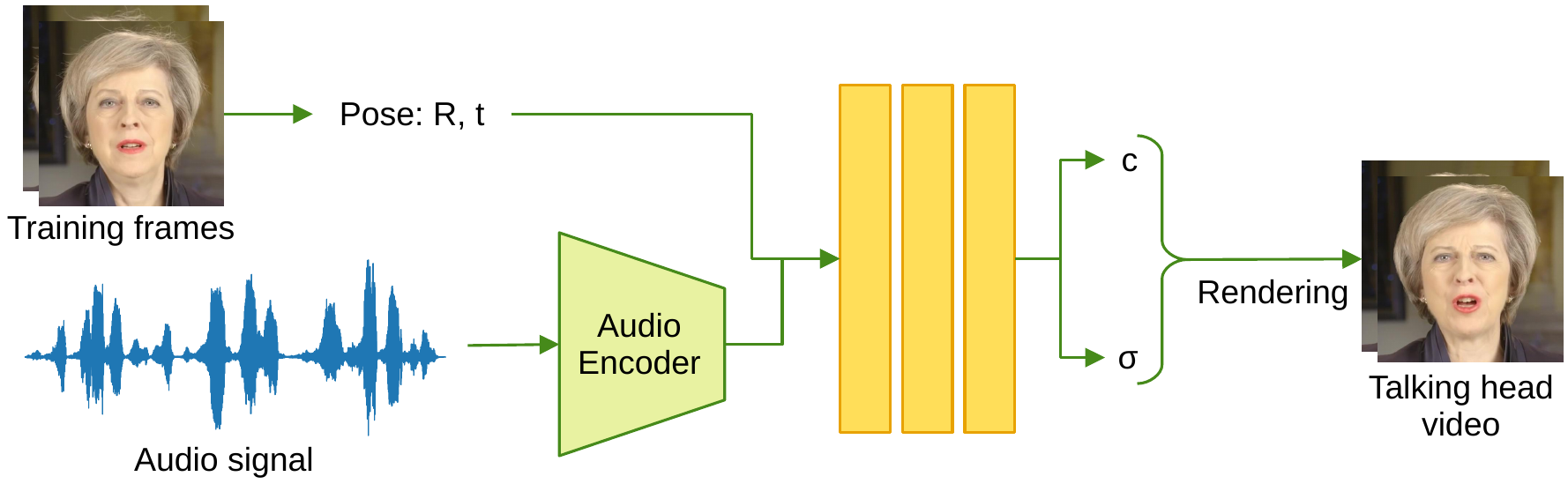}}
\vspace{-0.4cm}
\caption{An overview of talking-head synthesis based on NeRFs. The model learns to predict color and density values, which furtherenable the rendering of the deepfake talking head video.}
\Description{An overview of talking-head synthesis based on NeRFs. The model learns to predict color and density values, which furtherenable the rendering of the deepfake talking head video.}
\label{fig_nerf_deepfake}
\vspace{-0.5cm}
\end{center}
\end{figure}

\subsection{NeRF Usage Example}

NeRFs are used for talking head synthesis, as illustrated in Figure Figure \ref{fig_nerf_deepfake}. The main idea is to use the audio as the driving signal. More precisely, along with the information about the pose, the neural network processes audio features from a pre-trained encoder to output the color and density values from which the rendering is possible.

\section{Facial Forgery Localization}
\label{sec_supp_B}

Besides the standard classification task, we next review several studies \cite{huang-TIFS-2022,peng-ICMLCS-2023,wang-Electronics-2025,jiang-dc-2025,lai-CCBR-2023,miao-arXiv-2023, Mazaheri-WACV-2022, Tantaru-WACV-2024, Smeu-WACV-2025, Huang-CVPR-2025, Hong-CVPR-2024, Zhang-MM-2024} that address a specific detection subtask, namely facial forgery localization, as it represents a promising direction in attaining explainable deepfake detection. Such methods utilize features from two different spaces: pixel \cite{huang-TIFS-2022,peng-ICMLCS-2023,wang-Electronics-2025,lai-CCBR-2023} or frequency domain \cite{miao-arXiv-2023,jiang-dc-2025}. Huang \etal~\cite{huang-TIFS-2022} observe that the upsampling layers in GANs introduce a fake texture. Thus, the authors try to estimate the difference between paired real-fake images using an encoder-decoder model, in which an attention mechanism using the parsed face map is introduced as well. Wang \etal~\cite{wang-Electronics-2025} use the features extracted from two different pre-trained networks (Densenet-201 and Xception), and integrate spatial and channel attention mechanisms to better capture manipulation artifacts. Different from other works, Miao \etal~\cite{miao-arXiv-2023} and Jiang \etal~\cite{jiang-dc-2025} find that the deepfake artifacts are accentuated in the frequency domain. As a result, they combine low-level latent features through attention layers from both the RGB and frequency space. Some recent works~\cite{Smeu-WACV-2025, Huang-CVPR-2025} explore the use of multimodal models for deepfake localization. Smeu \etal~\cite{Smeu-WACV-2025} propose a convolutional decoder that leverages frozen CLIP embeddings to generate deepfake masks. Huang \etal~\cite{Huang-CVPR-2025} employ a multimodal large language model, introducing two novel tokens into the vocabulary to extract segmentation information. These tokens, combined with the visual features of the input image, are processed through a decoder to produce the final deepfake mask.

\section{Extended Comparative Analysis of Fusion Strategies for Multimodal Detectors}
\label{sec_supp_C}

We present the performance levels reported for various fusion strategies in Table \ref{tab:fusion_performance}. We only include methods that report the AUC for cross-dataset evaluation, which enables us to conduct a more robust analysis. Late fusion seems to have a competitive edge over early fusion, showing greater potential. This probably explains why late fusion received more attention in research studies. Salvi \etal~\cite{Salvi-JI-2023} explicitly explore the three different fusion strategies of multimodal models. We display their comparative study in Table \ref{tab:fusion_performance_salvi}. By examining the performance of all three strategies, we observe that early fusion achieves the best performance. Salvi \etal~\cite{Salvi-JI-2023} rationalize that the relationships between modalities are better extracted and understood when analyzing their interaction as early as possible. In summary, we find that distinct fusion strategies exhibit different advantages, and the choice of method should be carefully considered in relation to the desired use case.

\begin{table}[t!]
    \centering
    \caption{Performance (in terms of AUC) of various multimodal deepfake detection methods, which are organized according to their fusion strategy. Cross-dataset results are reported on FakeAVCeleb \cite{khalid-NeurIPS-2021}, DFDC \cite{dolhansky-arXiv-2020}, and KoDF \cite{Kwon-ICCV-2021}. The DFDC test set is different from one method to another, as there is no official split.}
    \vspace{-0.3cm}
    \begin{tabular}{lcccc}
    \toprule
         Fusion strategy & Paper & FakeAVCeleb & DFDC & KoDF\\
         \midrule
         \multirow{2}{*}{{Early}}
         & \cite{Nie-ACMMM-2024} & 0.931 & 0.854 & 0.935 \\
         & \cite{Salvi-JI-2023} & 0.900 & 0.810 & - \\
         \midrule
         \multirow{2}{*}{{Mid}}
         & \cite{Guo-CVPR-2025} & - & 0.877 & - \\
         & \cite{Yang-TIFS-2023b} & 0.828 & 0.806 & - \\ 
         \midrule
         \multirow{8}{*}{{Late}}
         & \cite{Cozzolino-CVPR-2023} & 0.941 & 0.952 & 0.899 \\
         & \cite{Feng-CVPR-2023} & - & - & 0.869 \\
         & \cite{Hosler-CVPR-2021} & - & 1.000 & - \\
         & \cite{Ilyas-ASC-2023} & 0.849 & - & - \\
         & \cite{Liu-SPIC-2023} & 0.864 & 0.841 & 0.853 \\
         & \cite{Miao-AAAI-2025} & - & - & 0.945 \\
         & \cite{Oorloff-CVPR-2024} & - & - & 0.955 \\
         & \cite{Smeu-CVPR-2025} & 0.946 & - & - \\
    \bottomrule
    \end{tabular}
    \label{tab:fusion_performance}
\end{table}

\begin{table}[t!]
    \centering
    \caption{Cross-dataset AUC reported by Salvi \etal~\cite{Salvi-JI-2023} for all fusing strategies on two popular datasets: FakeAVCeleb \cite{khalid-NeurIPS-2021} and DFDC \cite{dolhansky-arXiv-2020}.}
    \vspace{-0.3cm}
    \begin{tabular}{lcc}
    \toprule
         Fusion strategy & FakeAVCeleb & DFDC \\
         \midrule
         Early & 0.900 & 0.810 \\
         Middle & 0.830 & 0.710 \\
         Late & 0.850 & 0.730 \\
        \bottomrule
    \end{tabular}
    \label{tab:fusion_performance_salvi}
\end{table}

\section{Results on Popular Benchmarks}
\label{sec_supp_D}

\begin{table}[t!]
    \centering
    \caption{Results of top scoring image deepfake detection methods on DFFD~\cite{dang-CVPR-2020}, DiffusionFace~\cite{Chen-ArXiv-2024d},  ForgeryNet~\cite{He-CVPR-2021}. All methods are based on CNN architectures, the most utilized backbone being EfficientNet-B4. It is important to note that most of these methods leverage features from the frequency domain, besides those obtained from the pixel value space.}
    \begin{tabular}{llcc}
    \toprule
         Dataset & Method & Accuracy & AUC \\
         \midrule
            
         \multirow{3}{*}{DFFD~\cite{dang-CVPR-2020}} 
            & BNext-M~\cite{Lanzino-CVPRW-2024} & 99.18\% & 0.9994 \\
            & VGG-16~\cite{dang-CVPR-2020} & - & 0.9967 \\
            & XceptionNet~\cite{dang-CVPR-2020} & - & 0.9964 \\
        \hline
         \multirow{7}{*}{DiffusionFace~\cite{Chen-ArXiv-2024d}} 
            & GramNet~\cite{Liu-CVPR-2020} & 62.60\% & 0.7250\\
            & GFF~\cite{Luo-CVPR-2021} & 61.10\% & 0.7250\\      
            & RECCE~\cite{Cao-CVPR-2022} & 64.40\% & 0.7130 \\
            & F3Net~\cite{Qian-ECCV-2020} & 59.80\% & 0.6960 \\
            & HDP~\cite{Sun-IJCV-2025} & 63.10\% & 0.7309 \\
            & HFI~\cite{Choi-ArXiv-2024} & - & 0.7310\\
            & AEROBLADE~\cite{Ricker-CVPR-2024} & - & 0.6775\\
        \hline
        \multirow{4}{*}{ForgeryNet~\cite{He-CVPR-2021}}
            & SNRFilters-Xception~\cite{Guarnera-CVPRW-2020} & 81.09\% & 0.9052 \\   
            & GramNet~\cite{Liu-CVPR-2020} & 80.89\%  & 0.9020 \\
            & F3Net~\cite{Qian-ECCV-2020}  & 80.86\% &  0.9015 \\
            & XceptionNet~\cite{verdoliva-ICCV-2019}& 80.78\%  & 0.9012 \\
    \bottomrule
    \end{tabular}
    \label{tab:image_methods}
\end{table}
\paragraph{Image.} Table~\ref{tab:image_methods} includes top accuracy and AUC scores on three commonly-used datasets of deepfake images, namely DFFD~\cite{dang-CVPR-2020}, DiffusionFace~\cite{Chen-ArXiv-2024d} and ForgeryNet~\cite{He-CVPR-2021}. The results suggest that the oldest dataset, DFFD, has become saturated due to advancements in recent deepfake detectors. In contrast, the most recent dataset, DiffusionFace, featuring faces generated by diffusion models, poses a significantly greater challenge for state-of-the-art detectors. This highlights the need for future developments in deepfake detection to effectively differentiate between genuine faces and those synthesized by diffusion models.

\paragraph{Video.} Table~\ref{tab:video_methods} provides the performance levels of a few of the most effective methods for video deepfake detection on three distinctive datasets: FaceForensics++~\cite{verdoliva-ICCV-2019}, DeepFake Detection Challenge (DFDC)~\cite{dolhansky-arXiv-2020} and Celeb-DF~\cite{li-CVPR-2020b}. The main metric in this area is the AUC, but we also report the accuracy, whenever it is available. All the included models are outstanding, most of them almost achieving flawless performance. This is not only true for the newest methods that employ the most recent trends (such as transformers), but also for the preceding ones, that solely utilize CNNs. Nevertheless, on the more difficult datasets (DFDC and Celeb-DF), it can be observed that ViT-based architectures are superior.

\paragraph{Audio.} In Table \ref{tab:audio_results}, we report the Equal Error Rate (EER) values of top audio deepfake detection methods on ASVspoof 2019-LA~\cite{wang-CSL-2020} and 
ASVspoof 2021-LA~\cite{yamagishi-ASVspoof-2021}, two of the most popular audio deepfake detection datasets. GNN-based methods achieve the lowest EER values on both datasets, demonstrating their effectiveness in the synthesized speech detection task.

\paragraph{Multimodal.} In Table \ref{tab:multimodal_results}, we present the performance levels of top deepfake detection methods on two popular multimodal datasets, namely FakeAVCeleb \cite{khalid-NeurIPS-2021} and DFDC \cite{dolhansky-arXiv-2020}. For each method, we report the performance in terms of accuracy and AUC. For the FakeAVCeleb dataset, we include results from six state-of-the-art detection methods, with FRADE~\cite{Nie-ACMMM-2024} and AVFF~\cite{Oorloff-CVPR-2024} achieving the highest accuracy rates, both at $98.60\%$. On the DFDC dataset, we include five methods, with FRADE and PVASS-MDD demonstrating top performance, attaining accuracy rates of $97.20\%$ and $96.30\%$, respectively. The reported results hint towards important advancements in multimodal methods, with recent methods nearly saturating the benchmarks.

\paragraph{Cross-Compression.} An important practical aspect of deepfake detectors is their robustness to different compression levels. The most popular benchmark that can be used to evaluate compression robustness is FaceForensics++~\cite{verdoliva-ICCV-2019}, which has three distinct compression levels: raw (C0), high-quality (C23), and low-quality (C40). A standard evaluation protocol is to train the model on the high-quality (C23) subset and test its performance on the low-quality (C40) subset. In Table~\ref{tab:cross_compression}, we present results of various deepfake detectors on this evaluation protocol. As expected, all methods are impacted by the compression level, but the sensitivity varies significantly across architectures. Specifically, image-level models, such as RECCE and UCF, showcase a substantial drop in accuracy, suggesting that the frame-level visual artifacts they rely on are easily erased by compression, being insufficient for robust generalization. Since the performance on FaceForensics++ is nearing saturation, we also report results on a more recent dataset, CelebDF++~\cite{Li-arXiv-2025}. The observations are similar for this dataset, and the more pronounced drop in performance further highlights the necessity of achieving robustness against compression.

\begin{table}[t!]
    \centering
    \caption{Results of top scoring video deepfake detection methods on FaceForensics++~\cite{verdoliva-ICCV-2019}, DFDC~\cite{dolhansky-arXiv-2020} and Celeb-DF~\cite{li-CVPR-2020b} datasets. Transformer-based architectures usually outperform CNNs. Methods that use audio-visual chunks as input are superior to those that process individual frames and then aggregate the resulting latent representations.}
    \begin{tabular}{llcc}
    \toprule
         Dataset & Method & Accuracy & AUC \\
         \midrule
         \multirow{9}{*}{FaceForensics++~\cite{verdoliva-ICCV-2019}} 
            & TALL++ \cite{xu-ICCV-2023} & 98.65\% & 0.9987 \\
            & LipForensics \cite{haliassos-CVPR-2021} & 98.90\% & 0.9970 \\
            & FADE \cite{tan-AAAI-2023} & 92.89\% & 0.9952 \\
            & M2TR \cite{Wang-ICMR-2022} & 97.93\% & 0.9951 \\
            & App.+Beh. \cite{agarwal-WIFS-2020} & 98.90\% & 0.9900 \\
            & RealForensics\cite{haliassos-CVPR-2022} & - & 0.9900 \\
            & SUR-LID~\cite{Cheng-CVPR-2025} & - & 0.9485 \\
            & HDP~\cite{Sun-IJCV-2025} & - & 0.9507 \\
            & M2F2-Det~\cite{Guo-CVPR-2025b} & 98.79\% & 0.9934\\

        \hline
            
         \multirow{8}{*}{DFDC~\cite{dolhansky-arXiv-2020}} 
            & Efficient ViT \cite{coccomini-ICIAP-2022} & - & 0.9510 \\
            & TALL++ \cite{xu-ICCV-2023} & - & 0.9068 \\
            & CNN Ensemble \cite{bonettini-ICPR-2021} & - & 0.8782 \\
            & RealForensics\cite{haliassos-CVPR-2022} & - & 0.7590 \\
            & LipForensics \cite{haliassos-CVPR-2021} & - & 0.7350 \\
            & M2F2-Det~\cite{Guo-CVPR-2025b} & - & 0.8780\\
            & FreqBlender~\cite{Li-NeurIPS-2024} & - & 0.8756 \\
            & LAA-Net~\cite{Nguyen-CVPR-2024} & - & 0.8694 \\
         \hline

         \multirow{6}{*}{Celeb-DF~\cite{li-CVPR-2020b}} 
            & App.+Beh. \cite{agarwal-WIFS-2020} & 98.50\% & 0.9900 \\
            & FInfer \cite{hu-AAAI-2022} & 90.47\% & 0.9330 \\
            & RealForensics\cite{haliassos-CVPR-2022} & - & 0.8690 \\
            & LipForensics \cite{haliassos-CVPR-2021} & - & 0.8240 \\
            & M2F2-Det~\cite{Guo-CVPR-2025b} & 98.98\% & 0.9992\\
            & TALL~\cite{xu-ICCV-2023} & 97.57\% & 0.9855\\ 


        \bottomrule
    \end{tabular}
    \label{tab:video_methods}
\end{table}

\begin{table}[t!]
    \centering
    \caption{Results of top scoring audio deepfake detection methods on the ASVspoof 2019-LA~\cite{wang-CSL-2020} and ASVspoof 2021-LA~\cite{yamagishi-ASVspoof-2021} datasets. The top-performing audio deepfake detection methods~\cite{Chen-ICASSP-2023, Jung-ICASSP-2022} are based on graph-neural networks (GNNs). GNNs operate on graphs where nodes represent temporal segments, modeling inter-dependencies between segments. This allows the detection of artifacts introduced by deepfake generation methods, which are often present across multiple time segments.}
    \begin{tabular}{llc}
    \toprule
         Dataset & Method & EER (\%) \\
         \midrule
         
        \multirow{6}{*}{ASVspoof 2019-LA~\cite{wang-CSL-2020}}
        & GCN~\cite{Chen-ICASSP-2023} & 0.58 \\
        & Rawformer~\cite{Liu-ICASSP-2023} & 0.59 \\
        & AASIST~\cite{Jung-ICASSP-2022} & 0.83 \\
        & RawGAT~\cite{Tak-ASVSPOOF-2021} & 1.06 \\
        & TO-RawNet~\cite{Wang-INTERSPEECH-2023} & 1.58 \\
        & TSSDNet~\cite{Hua-SPL-2021} & 1.64 \\
        \hline

        \multirow{5}{*}{ASVspoof 2021-LA~\cite{yamagishi-ASVspoof-2021}} 
        & wav2vec2+AASIST~\cite{tak-ArXiv-2022} & 0.82 \\
        & wav2vec2+MLP~\cite{Martin-ICASSP-2022} & 3.54 \\
        & TO-RawNet~\cite{Wang-INTERSPEECH-2023} & 3.70 \\
        & Rawformer~\cite{Liu-ICASSP-2023} & 4.53 \\
        & AASIST~\cite{Jung-ICASSP-2022} & 9.15 \\
        \bottomrule

    \end{tabular}
    \label{tab:audio_results}
\end{table}

\begin{table}[t!]
    \centering
    \caption{Results of top scoring multimodal deepfake detection methods on the FakeAVCeleb~\cite{khalid-NeurIPS-2021} and DFDC~\cite{dolhansky-arXiv-2020} datasets. The top-performing methods~\cite{Miao-AAAI-2025, Nie-ACMMM-2024, Oorloff-CVPR-2024} are based on supervised learning. However, there are some unsupervised approaches~\cite{Smeu-CVPR-2025, Feng-CVPR-2023} that are trained exclusively on normal data and still achieve competitive results. A key advantage of these unsupervised methods is their typically superior generalization to novel deepfake techniques.}
    \begin{tabular}{llcc}
    \toprule
         Dataset & Method & Accuracy & AUC \\
         \midrule
         
        \multirow{8}{*}{FakeAVCeleb~\cite{khalid-NeurIPS-2021}} 
        & FRADE~\cite{Nie-ACMMM-2024} & 98.60\% & 0.9980 \\
        & AVFF~\cite{Oorloff-CVPR-2024} & 98.60\% & 0.9910 \\
        & MIS-AVoiDD~\cite{Katamneni-ICMLA-2023} & 96.20\% & 0.9730 \\
        & PVASS-MDD~\cite{Yu-TCSVT-2023} & 95.70\% & 0.9730 \\
        & SSVF~\cite{Feng-CVPR-2023} & 94.20\% & 0.9450 \\
        & MRDF~\cite{Zou-ICASSP-2024} & 94.05\% & 0.9243 \\
        & AVH-Align~\cite{Smeu-CVPR-2025} & - & 0.9460 \\
        & Multi-task AV~\cite{Miao-AAAI-2025} & 99.84\% & 0.9998 \\

 
        \hline

        \multirow{5}{*}{DFDC~\cite{dolhansky-arXiv-2020}} 
        & FRADE~\cite{Nie-ACMMM-2024} & 97.20\% & 0.9900 \\
        & PVASS-MDD~\cite{Yu-TCSVT-2023} & 96.30\% & 0.9890 \\
        & AVoiD-DF~\cite{Yang-TIFS-2023b} & 91.40\% & 0.9480 \\
        & AVA-CL~\cite{Zhang-ACM-2024} & 84.20\% & 0.8864 \\
        & AVFakeNet~\cite{Ilyas-ASC-2023} &  82.80\% & 0.8620 \\

        \bottomrule
    \end{tabular}
    \label{tab:multimodal_results}
\end{table}

\begin{table}[t!]
    \centering
    \caption{Cross-compression results of deepfake detection methods on FaceForensics++~\cite{verdoliva-ICCV-2019} and Celeb-DF++~\cite{Li-arXiv-2025}.}
    \begin{tabular}{llcccc}
    \toprule
         Dataset & Method & \multicolumn{2}{c}{High quality} & \multicolumn{2}{c}{Low quality} \\
         & & Accuracy & AUC & Accuracy & AUC \\
         \midrule
         \multirow{6}{*}{FaceForensics++~\cite{verdoliva-ICCV-2019}} 
            & TALL++ \cite{xu-ICCV-2023} & 98.65\% & 0.9987 & 92.82\% & 0.9457 \\
            & LipForensics \cite{haliassos-CVPR-2021} & 98.90\% & 0.9970 & 94.20\% & 0.9810 \\
            & FADE \cite{tan-AAAI-2023} & 92.89\% & 0.9952 & - & 0.8464 \\
            & Xception~\cite{dang-CVPR-2020} & 95.73\% & 0.9630 & 86.86\% &  0.8930 \\
            & RECCE~\cite{Cao-CVPR-2022} & 97.06\% & 0.9932 & - & 0.8190 \\
            & UCF~\cite{Yan-ICCV-2023} & - & 0.9705& - &	0.8399\\
        \hline
        \multirow{4}{*} {Celeb-DF++~\cite{Li-arXiv-2025}}
        & Xception~\cite{dang-CVPR-2020} & - & 0.7230 & - & 0.6480 \\
        & UCF~\cite{Yan-ICCV-2023} & - & 0.6850 & - &	0.5790 \\
        & IID~\cite{Huang-CVPR-2023} & - & 0.7050 & - & 0.5890 \\
        & Effort~\cite{yan-ICML-2025} & - & 0.8440 & - & 0.7920 \\
        
        \bottomrule
    \end{tabular}
    \label{tab:cross_compression}
\end{table}

\begin{figure}[!t]
\begin{center}
\centerline{\includegraphics[width=0.65\linewidth]{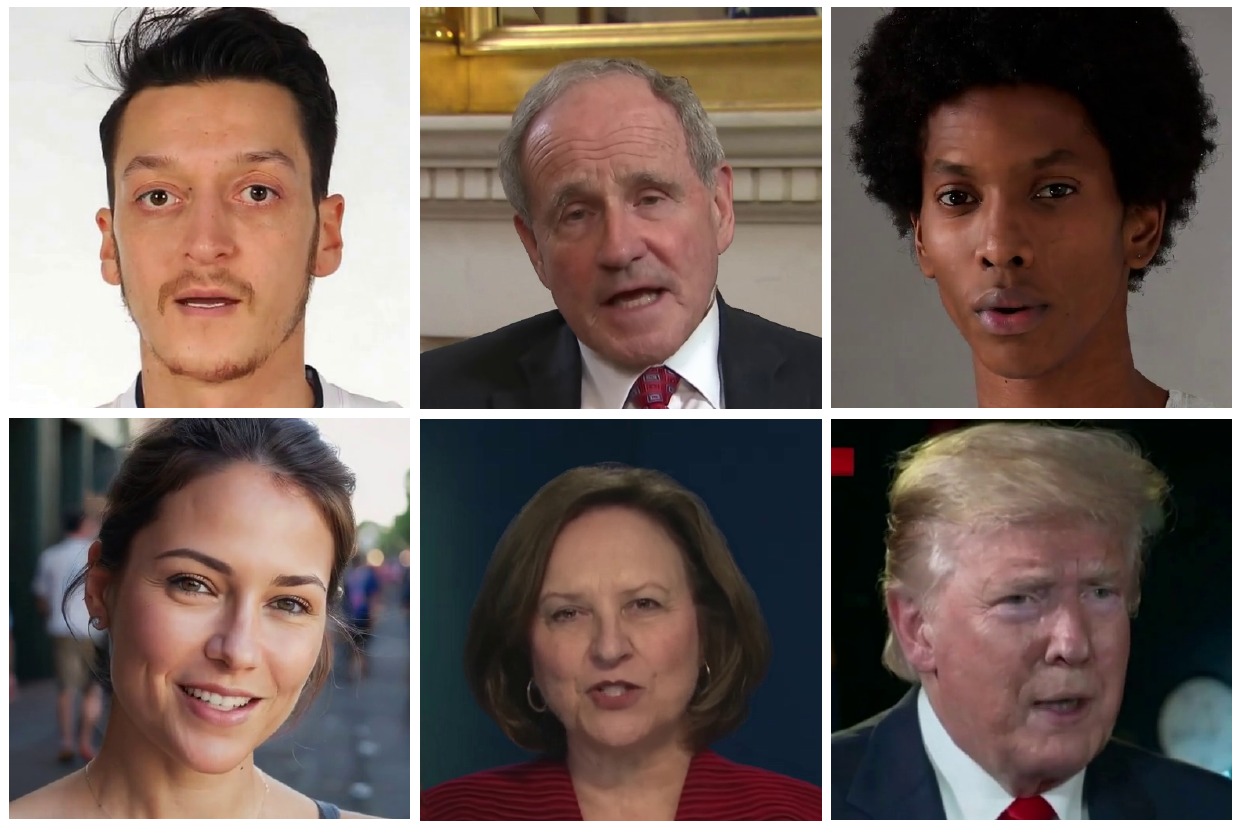}}
\vspace{-0.3cm}
\caption{Randomly sampled frames captured from the fake videos included in BioDeepAV exhibit a high level of realism. Best viewed in color.}
\Description{TODO}
\label{fig_samples_biodeep}
\vspace{-0.5cm}
\end{center}
\end{figure}

\begin{table}[t!]
    \centering
    \caption{Performance (in terms of average precision) of state-of-the-art detectors across different generative methods included in BioDeepAV. MRDF is trained on a distinct dataset than the other methods.}
    \vspace{-0.3cm}
    \begin{tabular}{lccccc}
    \toprule
        \multirow{2}{*}{Method} &  \multirow{2}{*}{Venue} & \multicolumn{4}{c}{Generative method} \\
        & & AniTalker & SyncTalk & GaussianTalker & EchoMimic \\
        \midrule
        XceptionNet~\cite{verdoliva-ICCV-2019} & ICCV 2019 & 0.4268 & 0.7837 & 0.5426 &0.4526\\
        F3Net~\cite{Qian-ECCV-2020} & ECCV 2020 & 0.3348 & 0.6462 & 0.4074 &0.4182\\
        RECCE~\cite{Cao-CVPR-2022} & CVPR 2022 & 0.3282  & 0.6319 & 0.5761 & 0.3850\\
        TALL~\cite{xu-ICCV-2023} & ICCV 2023 & 0.3261 & 0.7459 & 0.4469 & 0.4185\\
        UCF~\cite{Yan-ICCV-2023} & ICCV 2023 & 0.3505 & 0.6812 &  0.5433 & 0.4175\\
        \hline
        MRDF~\cite{Zou-ICASSP-2024} & ICASSP 2024  & 0.6146  & 0.6971 &  0.4545 &   0.6760 \\
        \bottomrule
    \end{tabular}
    \label{tab:benchmark results AP}
        \vspace{-0.3cm}
\end{table}

\section{Detailed Results on BioDeepAV}
\label{sec_supp_E}

We analyze the performance of several deepfake detectors for each generative method included in BioDeepAV. In Table~\ref{tab:benchmark results AP}, we present the average precision (AP) scores of the deepfake detectors for the fake class. An interesting observation is the relatively high performance of image-level methods on SyncTalk, suggesting that visual artifacts still provide significant cues for detection, in some cases. However, the better robustness of the multimodal MRDF approach underscores the limitations of unimodal detection, indicating that future research must look beyond visual features and take into account the synchronization between audio and visual streams.

\end{document}